\documentclass{article}

\usepackage{arxiv}

\usepackage[utf8]{inputenc} 
\usepackage[T1]{fontenc}    
\usepackage{hyperref}       
\usepackage{url}            
\usepackage{booktabs}       
\usepackage{amsfonts}       
\usepackage{nicefrac}       
\usepackage{microtype}      
\usepackage{lipsum}
\usepackage{graphicx}
\graphicspath{ {./images/} }
\usepackage{amsmath} 

\usepackage{soul}
\usepackage{caption}
\usepackage{tabularx}

\usepackage[T1]{fontenc}    
\usepackage{adjustbox}
\usepackage{subcaption}  
\usepackage{helvet}
\usepackage{tcolorbox}
\usepackage{multirow}

\usepackage[T1]{fontenc}
\usepackage{lmodern} 
\usepackage[utf8]{inputenc}
\usepackage{tikz}
\usetikzlibrary{arrows.meta, positioning, shapes.geometric, calc, shadows.blur}

\usepackage{tikz}
\usetikzlibrary{arrows.meta,positioning,calc}

\definecolor{aigold}{RGB}{244,210, 1} 
\definecolor{aigreen}{RGB}{210,244,211} 

\sethlcolor{aigreen}

\definecolor{aired}{RGB}{255,180,181}

\definecolor{aigold}{RGB}{255,180,181}

\definecolor{aiblue}{RGB}{173,216,230} 
 
\definecolor{lightred}{rgb}{1,0.9,0.9} 

\title{\textbf{Selective Imperfection as a Generative Framework for Analysis, Creativity and Discovery}}

\author{ \href{https://orcid.org/0000-0002-4173-9659}{\includegraphics[scale=0.06]{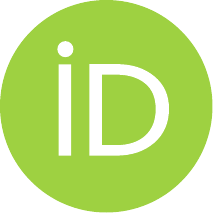}\hspace{1mm}Markus J. Buehler}\thanks{Corresponding author.} \\
	Laboratory for Atomistic and Molecular Mechanics\\Center for Computational Science and Engineering \\
        Schwarzman College of Computing \\
	Massachusetts Institute of Technology\\
	Cambridge, MA 02139, USA \\
        \\
	\texttt{mbuehler@MIT.EDU} 
}

\begin{document}
\maketitle


\begin{abstract}
We introduce materiomusic as a generative framework linking the hierarchical structures of matter with the compositional logic of music. Across proteins, spider webs, and flame dynamics, vibrational and architectural principles recur as tonal hierarchies, harmonic progressions, and long-range musical form. Using reversible mappings, from molecular spectra to musical tones and from three-dimensional networks to playable instruments, we show how sound functions as a scientific probe, an epistemic inversion where listening becomes a mode of seeing and musical composition becomes a blueprint for matter. These mappings excavate deep time: patterns originating in femtosecond molecular vibrations or billion-year evolutionary histories become audible. We posit that novelty in science and art emerges when constraints cannot be satisfied within existing degrees of freedom, forcing expansion of the space of viable configurations. Selective imperfection provides the mechanism restoring balance between coherence and adaptability. Quantitative support comes from exhaustive enumeration of all $2^{12}$ musical scales, revealing that culturally significant systems cluster in a mid-entropy, mid-defect corridor, directly paralleling the Hall-Petch optimum where intermediate defect densities maximize material strength. Iterating these mappings creates productive collisions between human creativity and physics, generating new information as musical structures encounter evolutionary constraints. We show how swarm-based AI models compose music exhibiting human-like structural signatures such as small-world connectivity, modular integration, long-range coherence, suggesting a route beyond interpolation toward invention. We show that science and art are generative acts of world-building under constraint, with vibration as a shared grammar organizing structure across scales. The framework supports multimodal \textit{de novo} protein design and AI systems that invent through collective dynamics.
\end{abstract}


\section{Introduction}
Science and art share a preoccupation with structure, pattern, and transformation. Matter and music each arise from simple elements arranged hierarchically: amino acids compose proteins that build resilient materials, while notes, motifs, and harmonic progressions cohere into forms that carry memory, surprise, and emotion. Across both domains, diverse function often emerges not from new building blocks but from reconfiguration across scales, and even from deliberate imperfection or broken symmetry \cite{Cranford2010NSA}. This perspective develops materiomusic as a practical framework (linking analysis, design, and creation) rather than a metaphor. We focus on bidirectional mappings between materials and sound, where physical structures (for example, three-dimensional spider webs) are translated to audio for exploration and performance, and where musical spaces serve as constructive representations for designing new molecular forms \cite{Su2018JRSi,Su2021CMJ,Su2022JMUI,Yu2019ACSNano,Qin2019EML}.
Another point of view that will be discussed here is that just as the first cave painters modeled the hunt through pigment and motion, the scientist today models the world through equations and experiments. Both acts emerge from the same generative drive -- that is, to construct, share, and test models of reality~\cite{Whitehead1929,Deleuze1994,Bergson1911}. Materiomusic situates itself within this lineage: a continuation of humanity’s oldest creative impulse, where art and science converge as complementary acts of world-making.

At its core, this article argues for materiomusic as a physically grounded, reversible generative framework linking matter, sound, and intelligence. Our central contribution is not a single experiment, but the demonstration that vibrations as preserving mappings can function as design operators across domains, enabling discovery rather than metaphorical translation.

We emphasize reversibility and structure preservation as the
defining features of materiomusic. The goal is not to ``score'' data, but to
construct mappings that mirror relationships across domains, an isomorphism where
possible, so that sequences, geometries, or fields translated into music can be
mapped back without losing internal logic. This bidirectionality enables
composition-as-design: musical operations (motivic recurrence, thematic recall,
long-range closure) become hypotheses about structure and function that can be
tested computationally and experimentally. Herein, the framework discussed in this paper redefines creativity as a physical, cognitive, and aesthetic process of resonance. Vibration operates as the universal medium through which matter, mind, and meaning co-generate structure, and selective imperfection serves as the algorithm that enables the universe to compose itself. Hence, vibration is both the structure and the computation of reality, and the use of symmetry breaking mechanisms  is its syntax.

Beyond structural mapping, materiomusic also excavates what we call deep time~\cite{Buehler2021CASTDeepTime}: patterns and memories of structure that once existed physically yet persist subliminally in creative output. By translating matter into sound and back, musical perception becomes a microscope that renders ancient or forgotten architectures audible - linking the molecular present to cultural and evolutionary pasts, and projecting them toward new forms.  
Across these examples, deep time is not only geological or cultural; it spans from femtosecond molecular motions to evolutionary histories (e.g. genetic evolution of DNA or proteins) and even cosmological origins. By coupling mappings between matter and music with AI, we can connect nano-scale vibrations, organismal behaviors, and cultural memory on a single perceptual plane, turning subliminal residues of structure into audible phenomena and actionable design cues.

Research over the past two decades has traced a trajectory from uncovering the molecular principles of protein-based materials, to translating these rules into engineered composites and membranes, and more recently to developing physics-aware, multi-agent AI frameworks that can automate aspects of discovery. Early studies of collagen and spider silk demonstrated how weak molecular bonds and hierarchical architectures generate extraordinary resilience, establishing transferable design rules that informed computation-to-fabrication approaches in bio-inspired composites and graphene assemblies \cite{Buehler2006Nature,Dimas2013AdvMater,Lu2023PNAS}. Building on this foundation, recent advances have integrated machine learning and multi-agent reasoning with mechanics, reframing AI from a predictive accelerator into a generative partner in science \cite{Yu2019ACSNano,Buehler2025ADGR,Buehler2025MusicSwarm}. In parallel, these ideas have extended into the cultural domain through materiomusic, where proteins, spider webs, and fractures have been translated into sound and composition. This body of work motivates the present perspective, which argues that matter and music share a generative language that can be harnessed for discovery, design, and artistic expression.

Concretely, advances in high-fidelity digitization of 3D webs and model-based sonification have enabled interactive instruments and installations that reveal latent architectural information through listening and performance \cite{Su2018JRSi,Su2021CMJ,Su2022JMUI}. At the molecular scale, a self-consistent sonification grounded in amino-acid vibrational physics provides a reversible map between sequences and music, supporting machine learning workflows that generate de novo proteins \cite{Yu2019ACSNano,Qin2019EML}. Extending beyond analysis, generative models have begun to capture and synthesize graph architectures of webs, illustrating how design rules can be learned and recomposed into novel structures \cite{Lu2023PNAS}.

\begin{figure}[ht]
\centering
\includegraphics[width=\textwidth]{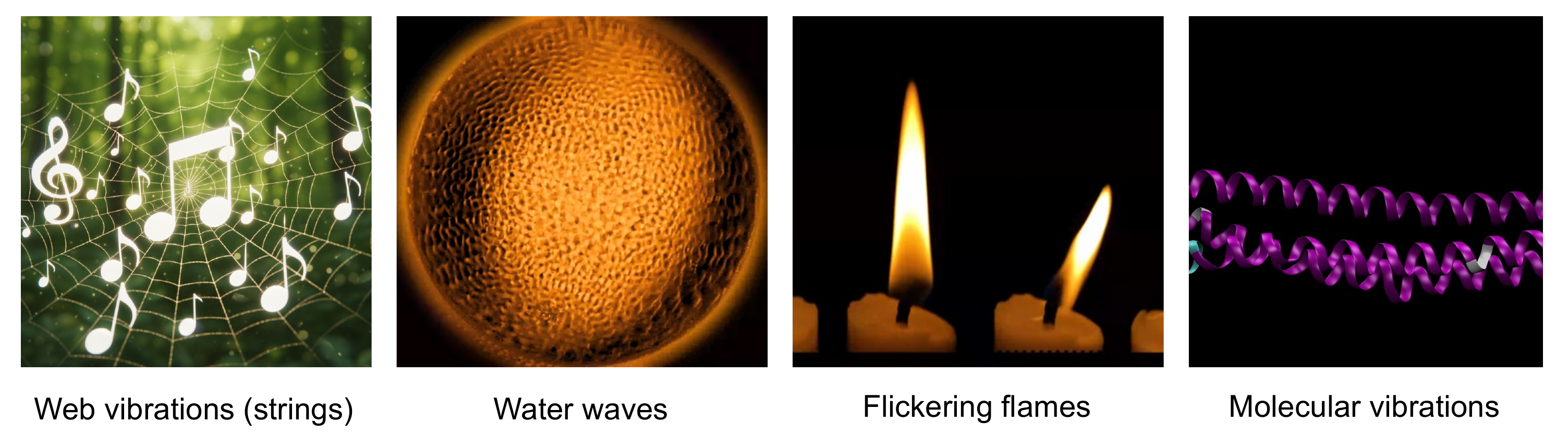}
\caption{Vibrations as generative algorithms of matter and music, physically 
grounded in the constraints of wave media. 
(A) Spider web vibrations behave like resonant strings, anchoring the spider’s 
perception in a vibrational universe defined by geometry and tension. 
(B) Water waves manifest oscillatory patterns that encode frequency, interference, 
and resonance, enforcing continuity and boundary conditions that regularize form. 
(C) Flickering flames exemplify vibrational instabilities in gases, producing 
rhythm and fluctuation governed by fluid and thermal dynamics. 
(D) Molecular vibrations, here along a protein backbone, reveal nanoscale 
oscillations dictated by atomic bonds that determine material and functional 
properties. 
Across these examples, oscillations are not arbitrary or symbolic but physically grounded: each medium acts as a co-author, embedding 
conservation laws and constraining the space of realizable patterns. 
Vibrations therefore serve not merely as physical phenomena but as generative 
algorithms, producing hierarchies and rhythms that can be read as both music and 
matter. In this sense, oscillation and resonance form a common computational 
substrate that simultaneously composes structure in the material world and 
organizes sound in the musical domain, providing a reversible and physically 
meaningful bridge between the two. Across media, vibration imposes physical constraints that make mappings invertible.}
\label{fig:vibration_generative}
\end{figure}

A central principle that is used in materiomusic is physical grounding. By this we mean 
that the mappings between matter and music are not imposed arbitrarily, but are 
anchored in the actual physics of vibrational media (Fig.~\ref{fig:vibration_generative}). A spider web vibrates only 
in modes permitted by its tension and geometry; a water surface forms standing 
waves constrained by continuity and resonance; a flame flickers according to the 
fluid and thermal dynamics of combustion; and proteins oscillate in normal modes 
dictated by their chemical bonds. In each case, the medium acts not merely as a 
passive canvas but as a co-author, enforcing conservation laws and 
boundary conditions that regulate the generative process. This differs sharply 
from symbolic sonification or superficial sound design, where relationships are 
layered onto data without preserving internal structure. Physical grounding 
ensures reversibility: the same mappings that translate sequences or geometries 
into music can be inverted to recover viable molecular or structural candidates. 
It also provides a natural regularization for artificial intelligence, since 
swarm agents that operate within physics-rich domains are constrained to produce 
outputs that are both inventive and realizable. In this sense, vibration and 
resonance are not metaphors but computational substrates, where they are shared grammars that 
govern the emergence of structure in both matter and music.

The article proceeds by outlining shared principles of structure and form across matter and music; presenting case studies in spider-web sonification, protein music, and cross-material mappings; and discussing how collective, agentic AI (for instance, swarm-based approaches to long-range musical form) can act as a creative partner that composes, not just interpolates. We close with implications for scientific discovery and engineering design, and a roadmap for how materiomusic can operate as a natural language for creativity that unifies molecules, materials, and sound.

\subsection{Foundational Ideas}
At its core, the materiomusic framework recognizes vibration as the universal mediator between matter and sound. Proteins, spider webs, and even fire all express themselves through oscillations that encode structure, function, and evolution across scales. When translated into the musical domain, these vibrational patterns become perceptible, allowing us to experience molecular dynamics or architectural hierarchies through listening. Conversely, musical composition can serve as a blueprint for matter: motifs and harmonic architectures map onto sequences and structural assemblies that can be synthesized and tested in the laboratory. This isomorphism reveals a profound symmetry: what emerges as resilience in silk, memory in proteins, or closure in symphonies is driven by the same principles of hierarchy, imperfection, and collective dynamics. By embedding these principles into agentic artificial intelligence systems, such as swarm-based models that evolve roles and weave long-range connections, we move beyond interpolation toward invention~\cite{Buehler2025ADGR, ghafarollahi2025sparksmultiagentartificialintelligence,Buehler2025MusicSwarm,wang2025swarmslargelanguagemodel}. Materiomusic thus not only bridges art and science but also points toward a generative paradigm of discovery, where sound becomes a microscope for matter and matter a score for creativity \cite{Su2021CMJ,Yu2019ACSNano,Lu2023PNAS,Buehler2025MusicSwarm}.

Materiomusic also reframes the notion of deep time. Traditionally used in 
geology to describe millions or billions of years, here it spans both 
ultrafast molecular vibrations on femtosecond scales and the cultural 
memory encoded in musical forms across centuries. By listening to proteins, 
webs, or flames, we access constraints and motifs that originate in ancient 
evolutionary processes, while at the same time projecting them into the 
domain of human culture via a universal medium of communication. This extension across temporal scales highlights 
materiomusic as a framework that bridges physics, biology, and cultural 
history through the shared grammar of vibration.

A central theme of this research has been to reveal how generative algorithms underlie the emergence of complexity in both natural and artificial systems. By studying proteins, spider webs, and other hierarchical materials, we have shown that simple rules of assembly, when orchestrated collectively, give rise to architectures of extraordinary resilience and adaptability. This insight extends to music, where motifs and harmonic constraints act as generative seeds that unfold into long-range form and emotional coherence. The work demonstrates that these generative principles are universal: they can be harnessed to design new proteins and materials through sonification, to compose music that mirrors biological organization, and to create agentic AI systems that discover by composing rather than interpolating. In this way, generative algorithms are elevated from abstract formalism to a tangible design language that unites matter, sound, and intelligence.

\subsection{Outline}

As an outline to this article, this perspective advances a single generative principle and develops its consequences across matter, music, and AI:

\begin{itemize}
  \item \textbf{Core mechanism: constraint-induced novelty:}
  Novelty arises when imposed constraints make the current representational space infeasible, forcing the introduction of a new degree of freedom (new variable, operator, or scale) that expands the space of viable configurations. This is a redefinition of creativity not as ``inspiration,'' but as constraint failure. 

  \item \textbf{Selective imperfection as a canonical route to novelty:}
  Defects, asymmetries, and heterogeneities are not noise; they are introduced degrees of freedom that enable systems to escape brittle optima and occupy a stable “sweet spot” balancing coherence and adaptability.

  \item \textbf{Why vibration as a lens for discovery:}
  Reversible, physics-grounded mappings between material dynamics and musical structure provide translation operators that (i) expose representational failure and (ii) make representational expansion actionable. The key insight is that vibration serves as a shared substrate linking structure, dynamics, and perception that allows us to develop deeper general insights.

  \item \textbf{Two isomorphic instantiations developed in the paper:}
  In matter, new degrees of freedom emerge as effective variables (interfaces, phase heterogeneity, defect distributions) that control robustness and energy dissipation; in music, new degrees of freedom appear as scale ``defects'', modulation, motif layering, and form-level operators that enable tension, resolution, and long-range closure.

  \item \textbf{Operationalization via agentic AI and validation:}
  Swarm/agentic generation in AI is framed as a mechanism to search for new degrees of freedom beyond interpolation, with success evaluated using structural signatures (e.g., small-worldness, modular integration, and long-range coherence used as proxy).
\end{itemize}

\section{Shared Principles of Structure and Form}

Across domains as diverse as materials, music, and intelligence, complex structure does not arise from isolated elements but from the self-organization of relationships across scales. These systems share common generative principles that govern how local interactions give rise to global form, coherence, and function. In the following, we identify such shared principles, beginning with hierarchy as a foundational mechanism linking structure, perception, and resilience.

\subsection{Hierarchies Across Scales}

Across both material and musical systems, structure emerges through nested hierarchies that connect the microscopic to the macroscopic. In biology, proteins exemplify this principle: amino acid sequences fold into secondary motifs, which organize into domains, fibers, and finally whole tissues. Each level of assembly adds functionality and resilience, with global properties that cannot be inferred from the local scale alone \cite{Cranford2010NSA}. Spider webs offer another striking example, where nanoscale silk proteins combine into fibrils and threads, which then assemble into the intricate geometry of the full web. This multiscale hierarchy enables extraordinary toughness, adaptability, and multifunctionality \cite{Su2018JRSi,Su2021CMJ}.

Music is structured in a similar way. Notes and intervals combine into motifs, which build into phrases, harmonies, and symphonies. Just as the resilience of spider silk depends on the interplay of scales, musical meaning and emotional power arise from relationships between levels: how small motifs recur, vary, and integrate across the whole piece. Hierarchical organization therefore provides a common design principle across domains: both proteins and spider webs, like motifs and symphonies, show how local building blocks gain richness and coherence when embedded in layered, multiscale structures \cite{Yu2019ACSNano}.

These physical hierarchies are mirrored in perception. For the spider, mechanoreceptors distributed across its legs form a decentralized sensorium; while for the human, auditory and motor cortices recursively encode rhythm and form. Both organisms construct coherence not from a central plan, but from hierarchical feedback across distributed sensors. Perception thus participates in the same layered architecture that defines matter and music.

\begin{figure}[ht]
\centering
\includegraphics[width=0.9\textwidth]{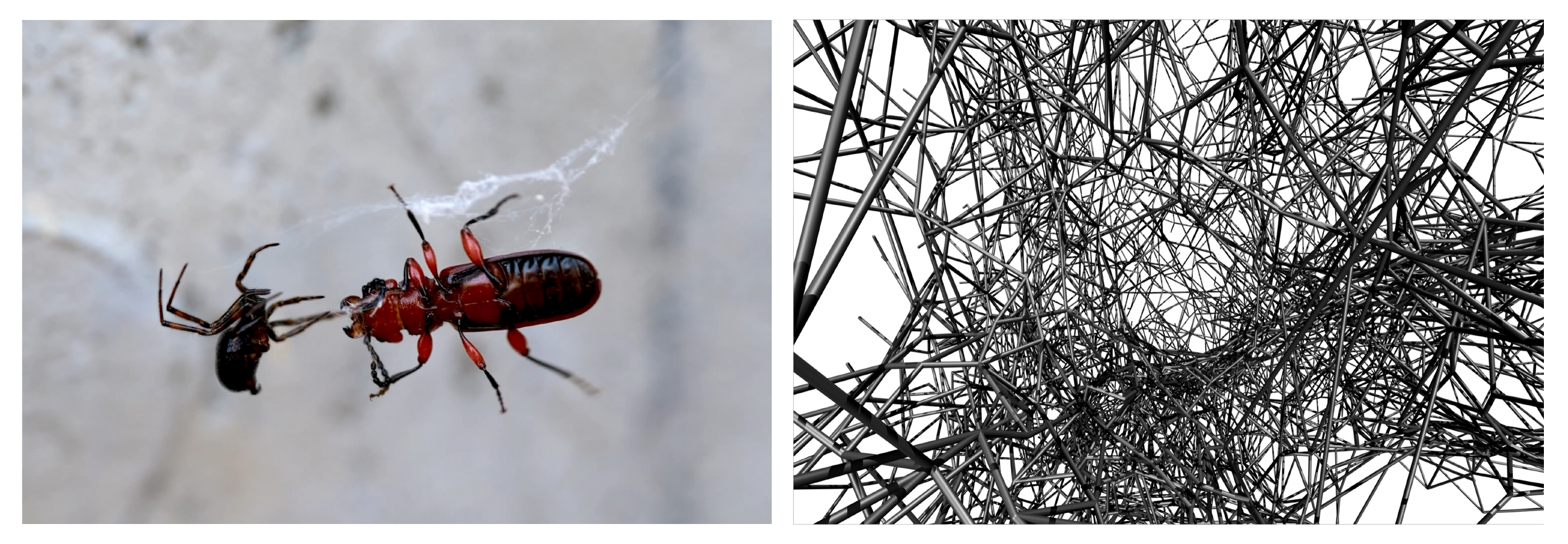}
\caption{Vibrational ecology of predation in a spider web.
Panel A: A false widow spider (\textit{Steatoda grossa}) subdues a cereal leaf beetle (\textit{Oulema melanopus}) in its cobweb. The spider, identified by its bulbous, glossy abdomen and uniform dark coloration, relies on vibrational cues transmitted through its silk to locate and envenomate the struggling prey. The web thus serves a dual function: a mechanical trap that restrains the beetle and a sensory organ that extends the spider’s perception into a vibrational universe. Panel B: A digital model of a 3D spider web~\cite{Su2018JRSi}.}
\label{fig:vibrational}
\end{figure}

Spiders such as \textit{Steatoda grossa} inhabit a world defined not primarily by vision, but by vibration. Their irregular cobwebs form a resonant network in which every prey movement, courtship signal, or environmental disturbance propagates as mechanical waves. In Fig.~\ref{fig:vibrational}, the captured cereal leaf beetle thrashes within the web, sending irregular vibrational signatures that the spider detects through specialized mechanoreceptors in its legs. This vibrational feedback enables the spider to distinguish prey from background noise, navigate toward it, and deliver immobilizing silk and venom. In this way, the spider’s silk functions simultaneously as a structural material and a sensory extension, embodying the broader principle that richness and resilience emerge from living in (and exploiting) a heterogeneous vibrational environment.

\subsection{Universality and Diversity}
\begin{figure}[h!]
    \centering
    \includegraphics[width=\textwidth]{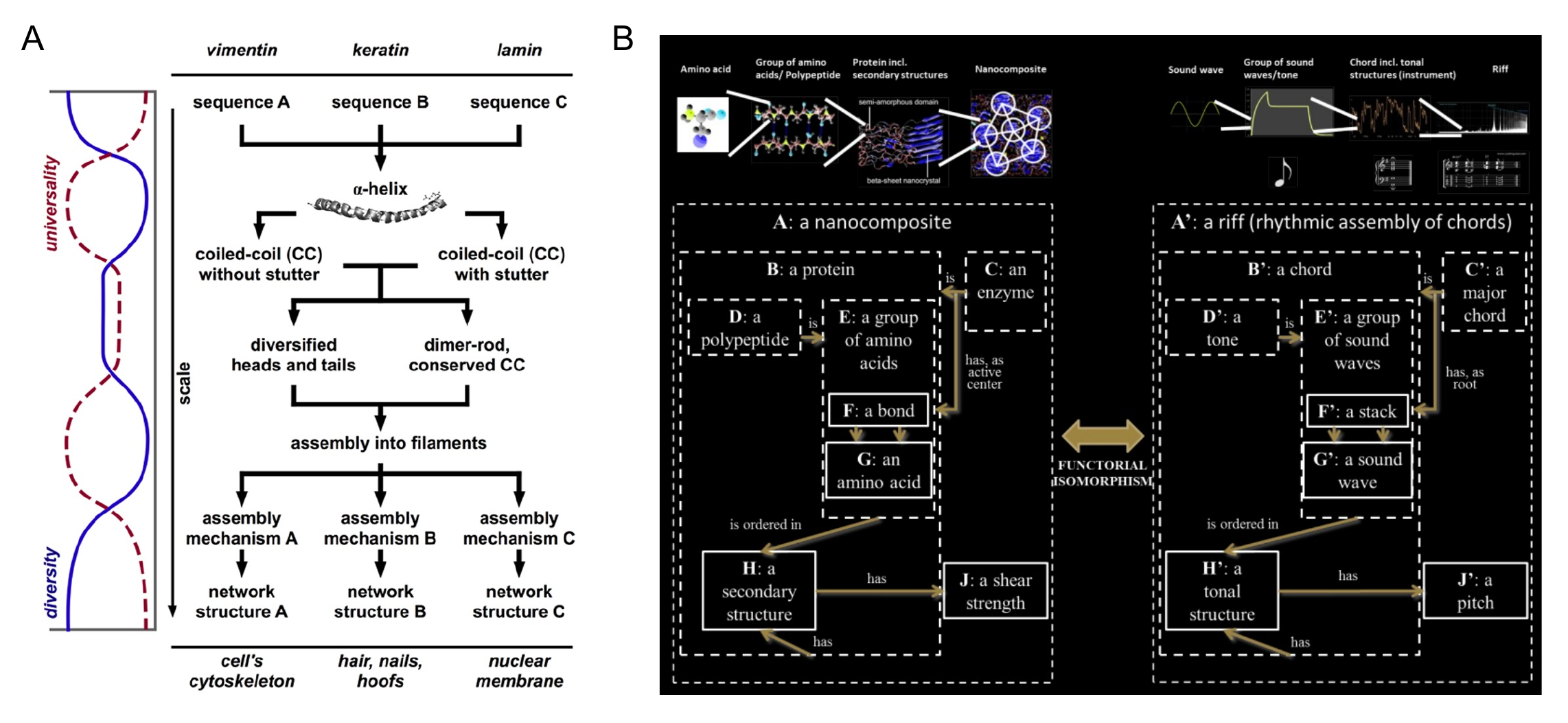}
    \caption{Functorial isomorphism between hierarchical protein materials and musical composition.
    (A) Hierarchical assembly of intermediate filament proteins (vimentin, keratin, and lamin). The diagram illustrates the progression from distinct amino acid sequences to coiled-coil dimers and tetramers, eventually forming complex filamentous networks with specific mechanical properties, representing the Univesality-Diversity-Principle (UDP)~\cite{Ackbarow2009Alpha-helicalFlaw-tolerant,Cranford2010NSA}. 
    (B) Category-theoretic mapping establishment as reported in~\cite{Giesa2011ReoccurringAnalogies}. A rigorous mathematical isomorphism is defined between the domain of matter (left) and music (right). Fundamental building blocks (amino acids) map to sound waves, polypeptides map to tones, and full proteins map to chords. This mapping extends to higher-order structures, where a material nanocomposite (A) corresponds to a musical riff (A'), and specific properties such as shear strength (J) translate to musical pitch (J'). This framework enables the direct translation of material assembly mechanisms into musical composition rules, and vice versa. Panel A is reprinted from~\cite{Cranford2010NSA}, and panel B from~\cite{Giesa2011ReoccurringAnalogies}.}
    \label{fig:materiomusic_isomorphism}
\end{figure}

A recurring theme across both matter and music is the balance between universality and diversity. In biological systems, nearly all proteins are constructed from the same twenty amino acids, yet the vast combinatorial space of sequences and folding pathways yields an astonishing diversity of structures and functions. Spider silk, collagen, keratin, and enzymes all arise from the same alphabet, differentiated not by building blocks but by hierarchical arrangement and subtle variation in motif usage \cite{Cranford2010NSA}. Universality provides the foundation (the alphabet and grammar) while diversity emerges through reconfiguration, defect incorporation, and evolutionary pressure (Fig.~\ref{fig:materiomusic_isomorphism}).

The same principle defines musical systems. Twelve tones per octave constitute the universal scaffold in Western music, but through selection of scales, rhythmic choices, and the creative bending of rules, composers generate infinite diversity of style and affect. Seven-note scales, for example, are particularly abundant because they balance structural coherence with the introduction of imperfections that create tension and resolution. Music thus demonstrates how universality does not limit creativity but enables it: rules provide recognizable structure, while variation and violation provide richness, novelty, and emotional depth.

This universality–diversity paradigm highlights a generative mechanism common to matter and music: complex, adaptive outcomes emerge not by inventing new fundamental components but by reconfiguring a small set of universal elements across scales. In materials, this yields multifunctionality and resilience; in music, it yields narrative, closure, and surprise. In both cases, simplicity at the foundation amplifies creative potential, demonstrating that diversity and novelty are themselves products of constraint \cite{Yu2019ACSNano,Su2021CMJ}.

Viewed through energy flow, music itself can be understood as a localization process: selecting and organizing spectral energy out of broadband noise into structured form. This continuum from white noise to tempered consonance to deliberate ``surprise'' mirrors how materials organize from randomness into hierarchical patterns, emphasizing that diversity arises not from new alphabets but from selective localization and recombination across scales.  

\begin{figure}[ht]
\centering
\includegraphics[width=0.9\textwidth]{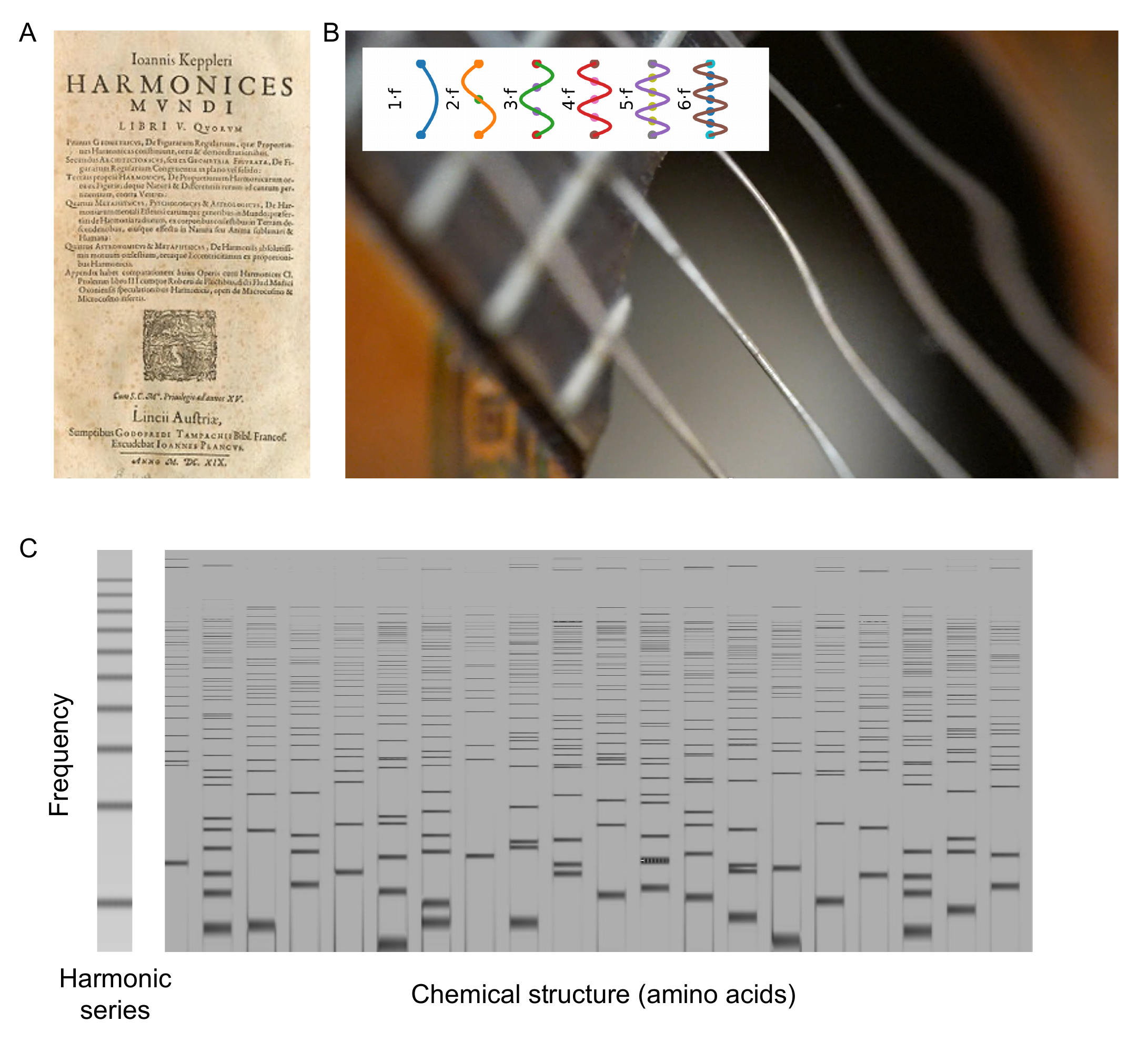}
\caption{A perspective that outlines the relationship from cosmic to molecular harmony, each touching on particular nuances of the nature of vibrational features that occur at distinct scales.
(A) Title page of Johannes Kepler’s \emph{Harmonices Mundi} (1619), which
sought cosmic order in planetary orbits expressed as musical ratios.
(B) Vibrating strings of a guitar, a physical realization of harmony as
resonance and frequency ratios. 
(C) Comparison of the melodic spectrum of the harmonic series (akin to panel B) with a representation of amino acid sequences: here the
chemical structure of proteins is translated into a notional ``scale''.
Unlike traditional harmonic scales, this mapping is not based on
integer ratios or consonance, but on structural encodings that give each
molecular building a unique pattern in sound~\cite{Yu2019ACSNano}. These
panels show how harmony, whether literal or metaphorical, provides
a framework for linking order across cosmology, acoustics, and biology. The comparison between the harmonic series and the properties of the amino acid scaling reveals distinct patterning of how sounds are generated through new constraints and rules.}
\label{fig:harmony}
\end{figure}

Figure~\ref{fig:harmony} highlights the enduring interplay of music, physics, and cosmology in light of the universality and diversity of building blocks. Kepler’s vision of a mathematically perfect cosmic harmony (panel A) reflects the aspiration toward universality and order, while the vibrating guitar strings (Fig.~\ref{fig:harmony}B) show harmony in its physical embodiment, as oscillations shaped by friction, material heterogeneity, and imperfection. In practice, it is precisely these irregularities that give sound its color and timbre, just as small deviations from ideal ratios enrich musical experience. The dialogue between abstract harmony and embodied vibration thus reinforces our central theme: richness and expressivity emerge not from idealized perfection, but from the subtle imperfections that shape real structures in matter and music alike.

\subsection{Imperfections as Richness}

While universality and ``harmony'' (Figure~\ref{fig:harmony}) provides the scaffolding for structure, our experience suggests that it is often imperfections that imbue both matter and music with resilience and expressive depth, but a greater mathematical analyzes is require to understand these ideas at a more foundational level. In materials science, defects play a dual role. Crystalline dislocations, vacancies, or grain boundaries disrupt perfect periodicity, but rather than simply weakening a material, they frequently enhance toughness and adaptability. For example, the controlled introduction of heterogeneities in composites can deflect cracks, dissipate energy, and improve resilience, turning flaws into features \cite{Buehler2006Nature,Dimas2013AdvMater}. Spider silk exemplifies this principle: nanoscale variations in the sequence and distribution of motifs create localized ``soft spots'' that prevent catastrophic fracture, enabling the web to combine strength with extensibility.

\begin{figure}[h!]
\centering
\includegraphics[width=1.\textwidth]{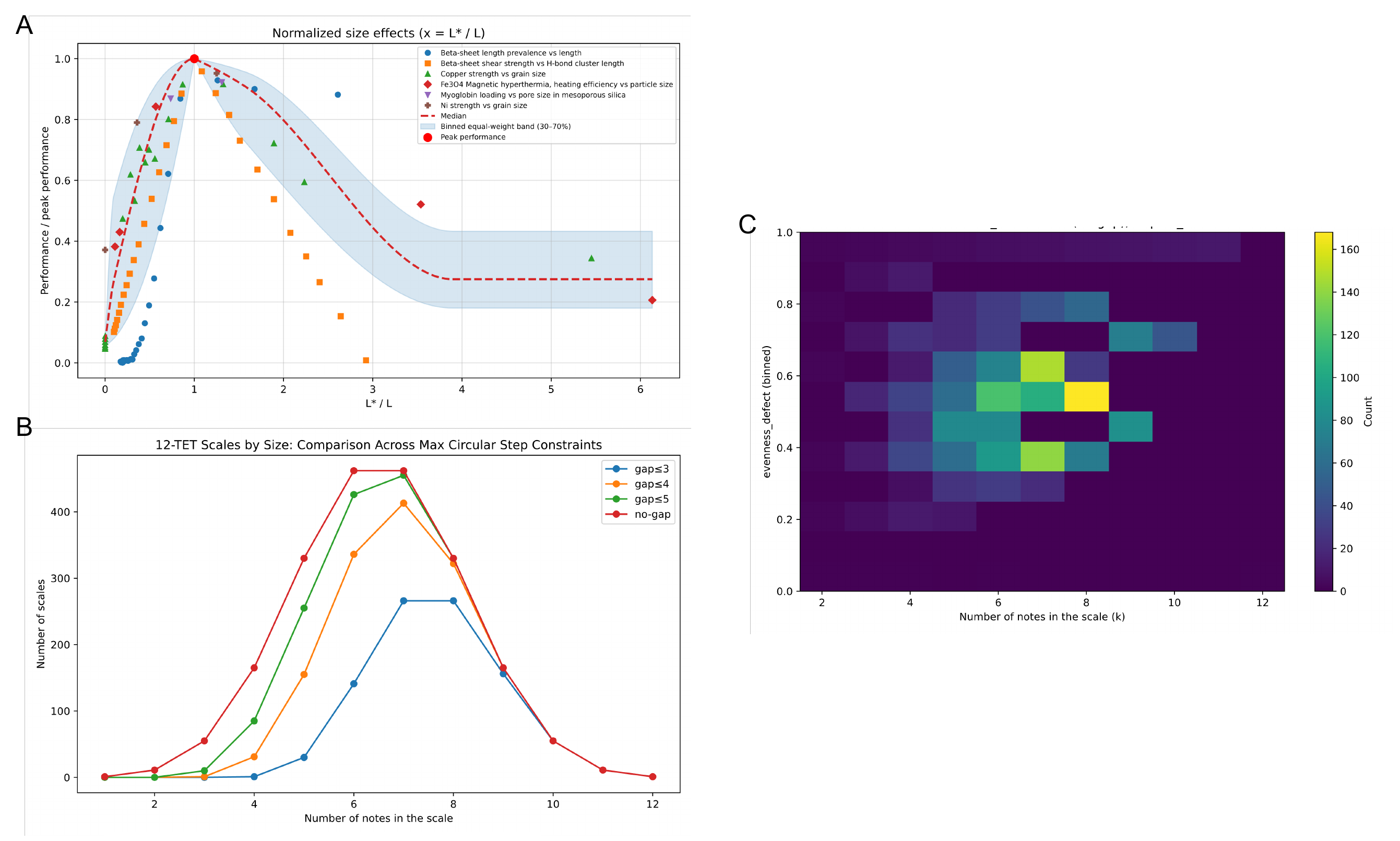}
\caption{Defects and scaling in matter and music.
(A) Normalized performance over inverse of a characteristic scale (featuring data from \cite{Armstrong2016HallPetch,Keten2008GeometricScalec,Miao2022SBA15Mb,Ma2004Fe3O4SAR,Stolyarov2020WearGrainSize} for copper, nickel, and biological systems including proteins). Smaller features correspond to larger $L^*/L$. In the case of grain size effects (Hall-Petch and inverse Hall-Petch), smaller grains (larger $L^*/L$) means
greater density of grain boundaries (potential defect sites). The data show
classical Hall-Petch strengthening at intermediate $1/L$, a peak at the critical
grain size $L^*$, and softening at very fine scales (large $1/L$)
(inverse Hall-Petch regime). 
(B) Enumeration of valid scales in 12-TET as a function of number of notes $k$,
under different maximum circular step constraints (max gap $\leq 3,4,5$, or no gap).
A valid scale is defined as a subset of the 12 pitch classes including the root
(note 0) and, if a gap constraint is applied, no adjacent interval exceeding
the threshold. Exhaustive enumeration ($2^{12}$ subsets) was performed by computing step vectors from bitmasks, filters by constraints,
and counts scales by size. 
(C) Distribution of evenness defect vs.\ number of notes $k$ (no gap constraint,
root required). The evenness defect is the normalized standard deviation of
step sizes relative to the uniform value $12/k$, rescaled to [0,1] by the
maximum observed standard deviation at that $k$. The heatmap shows scale counts
per bin (20 bins along the vertical axis). The concentration of scales around
moderate evenness ($\sim$0.4–0.6) and mid-size $k$ mirrors the combinatorial
dominance of 6–8 note scales. Together, these panels highlight a unifying
principle: both material strength and musical richness often emerge from intermediate
levels of imperfection.
In both matter and music we find scaling laws with a single optimum: materials exhibit maximum strength or function at a critical size, and pitch-class systems concentrate their diversity and balance at intermediate cardinalities. The alignment of these peaked behaviors suggests that the principle of a scale-dependent sweet spot is a universal feature of complex systems.
}
\label{fig:defects}
\end{figure}

As shown in Fig.~\ref{fig:defects}A, performance increases with inverse length scale $1/L$ until a critical point $L^*$, after which performance decreases. In the case of polycrystalline materials (copper, nickel as shown in the plot) this corresponds to the nanocrystalline regime. Therein, there exists a non-monotonic Hall–Petch behavior that highlights how an optimal density of grain boundaries provides the right balance of defect sources and barriers to maximize strength, while too many boundaries eventually lead to softening.

Music is likewise defined by its departures from perfect symmetry. Equal temperament,  a tuning system that divides the octave into twelve equal semitone intervals, represents a universal scaffold. Yet the most widely used scales, such as the seven-note diatonic modes, are imperfect subsets of this system: they omit certain pitches and introduce asymmetries that break the uniformity of the chromatic scale. Statistical analyses show that scales with three or four imperfections (that is, departures from full chromatic symmetry) dominate across cultures, suggesting that musical creativity thrives in the space between order and disorder \cite{AllTheScales,RingScales,Savage2015PNAS}. It is precisely these gaps, tensions, and asymmetries that allow for tonal hierarchy, expectation, and resolution, without which music would collapse into monotony or noise. 

As illustrated in Fig.~\ref{fig:defects}B--C, the exhaustive enumeration of all $2^{12}$ possible 12-TET subsets reveals that the combinatorial weight of the scale universe concentrates in the mid-range: scales with 6–8 notes dominate across different gap constraints (panel B). When analyzed by their evenness defect, these same mid-sized scales also cluster around moderate irregularity values of $\sim$0.4–0.6 (panel C). This means that most scales are neither perfectly uniform (like whole-tone or chromatic) nor maximally uneven, but instead occupy an intermediate space where balance and asymmetry coexist. Just as materials achieve optimal toughness at intermediate levels of defect density, musical variety and expressivity are enriched by this ``sweet spot'' of moderate imperfection in scale structure.

In direct analogy, grain size $d$ in a polycrystal sets the density of grain boundaries (smaller $d$ $\rightarrow$ more potential defect sites), just as the number of notes $k$ in a scale sets the number of intervals per octave (larger $k$ $\rightarrow$ more opportunities for asymmetry). Thus, $d^{-1}$ and $k$ play equivalent roles as measures of the \emph{capacity for imperfection}.
Hence, a unifying principle across materials and music is that selective
imperfection is generative. In metals, increasing grain-boundary density
strengthens a material up to a limit (Hall--Petch), beyond which the inverse
regime softens it. In 12-TET, a census of all chromatic subsets shows that
mid-sized scales (6--8 notes) with moderate unevenness dominate: perfectly
even sets tend to feel inert; highly uneven sets resist organization. The richest
behaviors arise in the middle where we see neither crystalline perfection nor chaotic
fragmentation, yielding toughness in matter and expressivity in music. 

Selective imperfection thus functions as a thermodynamic engine of creativity. It uses the strict constraints of one domain (whether physical stress or harmonic rules) to force the emergence of novelty in another, ensuring that the resulting structures are not arbitrary, but necessary solutions to a stability problem.

Indeed, the study of scale structure and pitch organization has long drawn on formal approaches, from Forte’s foundational set-theoretic analysis of atonal music \cite{Forte1973}, Rahn’s integer model of pitch \cite{Rahn1980}, and Hanson’s theory of harmonic materials \cite{Hanson1960}, to more recent explorations of pitch-class sets and computational tools \cite{NelsonPCSets,DuncanCMT}. Extensions into mathematical and combinatorial domains reveal how imperfections, circular distributions, and maximal area sets underlie musical variety and perception \cite{Bridges2005Circular,Rappoport2007,ZeitlerAllScales}.

\begin{figure}[h!]
\centering
\includegraphics[width=0.55\textwidth]{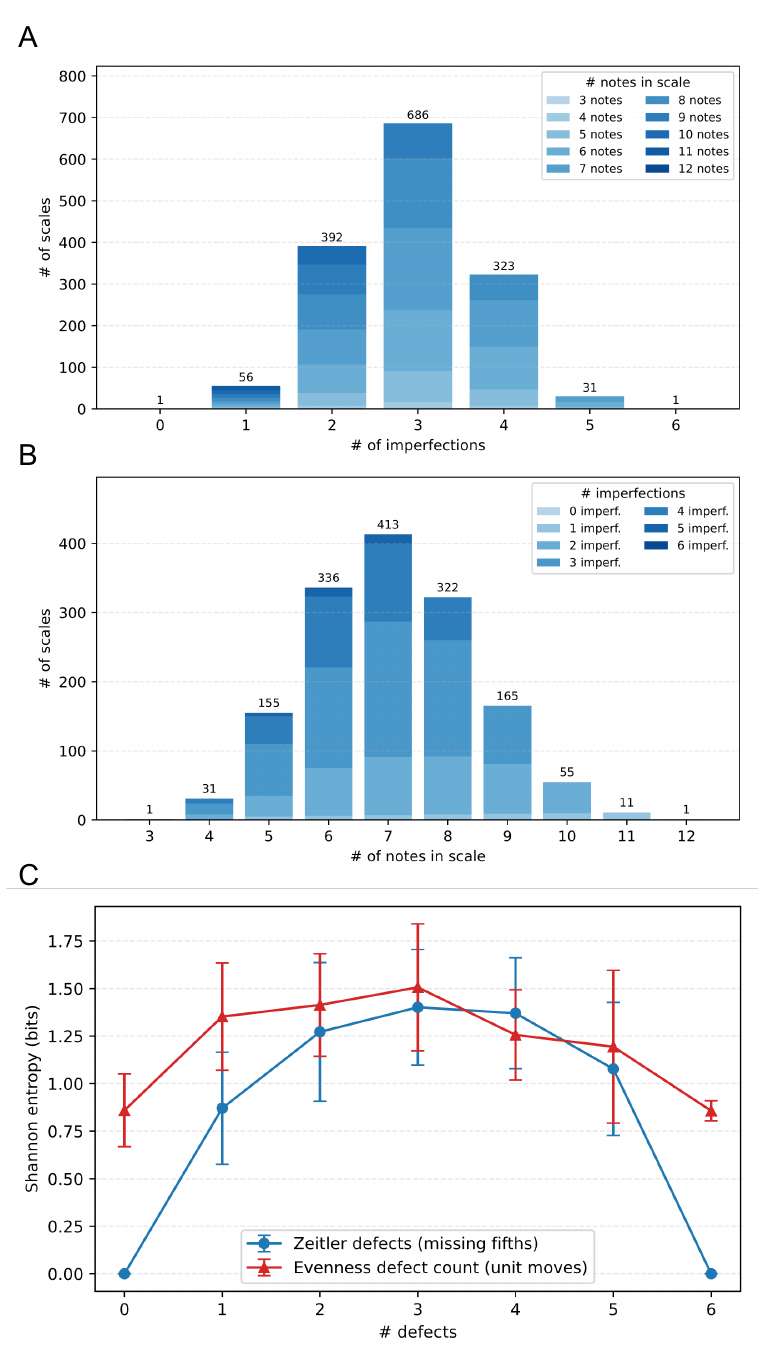}
\caption{ Combinatorial structure of valid $12$-TET scales as a function of defect number and interval entropy, for Zeitler's definition of defect. Panel A: Number of distinct scales as a function of the number of imperfections (defects). Each bar at a given defect count $k$ is stacked by the number of notes in the scale $n$ (from $3$ to $12$), revealing that valid scales are overwhelmingly concentrated in an intermediate defect regime ($k \approx 2$--$4$), while both perfectly self-contained scales ($k=0$) and highly defective scales ($k\geq 5$) are rare.
Panel B: Number of distinct scales as a function of the number of notes $n$ in the scale, with each bar stacked by defect number $k$. The distribution peaks at $6$--$8$ notes per scale and $2$--$4$ defects, showing that commonly sized scales tend to inhabit a middle band where harmonic connectivity (presence of internal fifths) and intervallic variety are jointly balanced.
Panel C: Mean Shannon entropy of the cyclic interval pattern as a function of defect number, with error bars indicating one population standard deviation (plotted for both Zeitler's definition of defect and the unevenness defect definition). In both measures, entropy rises from the low-defect regime to a broad maximum at $k=3..4$ and then declines again for very high defect counts. The results indicate that moderate loss of internal fifths is associated with the greatest diversity and irregularity of step sizes.
}
\label{fig:scales-defects-entropy}
\end{figure}

As another metric of a ``defect'' we use Zeitler's definition where we define the number of defects as the count of degrees whose internal perfect fifth is missing. This can be considered a functional defect from the perspective of tonal harmony (as opposed to the ``geometric defect'' or structural defect considered above).
Figure~\ref{fig:scales-defects-entropy} highlights a tight coupling between combinatorial abundance, harmonic connectivity, and intervallic complexity. Figure~\ref{fig:scales-defects-entropy}A Figure~\ref{fig:scales-defects-entropy}B show that, under our constraints, valid scales concentrate in a narrow region of parameter space: most scales have $6$–$8$ notes and $2$–$4$ defects, while both defect-free scales and those with many missing fifths are comparatively rare. Figure~\ref{fig:scales-defects-entropy}C reveals that this intermediate defect band also maximizes the Shannon entropy of the interval pattern, that is, scales with $2$–$4$ imperfections exhibit the richest and least predictable mixtures of step sizes, whereas very low or very high defect counts are associated with more regular or more constrained interval structures. Together, these results suggest that musically interesting scales may naturally arise in a regime where the scale is neither perfectly self-contained nor highly fragmented in terms of its internal fifth relations, but instead balances harmonic completeness and structural diversity to achieve both internal coherence and expressive flexibility.

\begin{figure}[htbp]
    \centering
    \includegraphics[width=\textwidth]{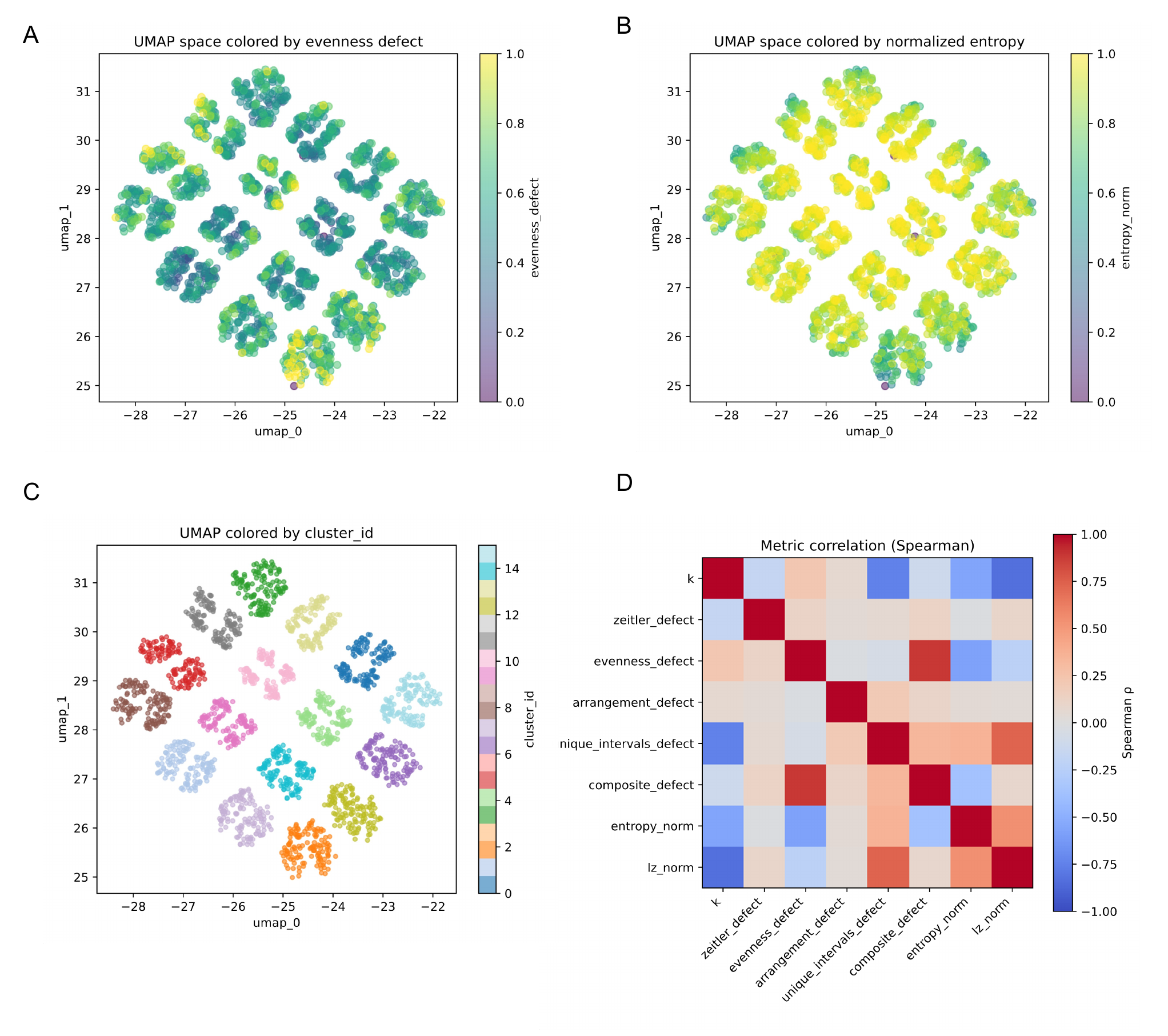}
    \caption{
     Geometric structure of the full 12-TET scale space and relationships between defect measures.
    Each point in panels A–C represents one of the $2^{11}=2047$ root-present $12$-TET pitch-class subsets (binary masks with pc~0 included), embedded with UMAP from the raw scale vectors. 
    Panel A: UMAP colored by evenness defect (normalized within each scale size $k$) reveals compact islands corresponding to combinatorial families of scales; within each island evenness varies smoothly in a mid-range, acting as a local order parameter rather than a single global axis.
    Panel B: UMAP colored by normalized Shannon entropy of the interval-size distribution shows entropy is high across most of the manifold, with only a few highly regular or very sparse patterns at the periphery exhibiting low entropy.
    Panel C: UMAP colored by $k$-means cluster identity ($k=16$) recovers well-separated islands that correspond to bands in scale size with distinct geometric-defect profiles.
    Panel D: Spearman correlation matrix between scale size, Zeitler defect, geometric defects (evenness, arrangement, unique-intervals, composite), normalized entropy, and LZ-normalized sequence complexity shows that functional defect is largely independent of geometric defect, while (as expected) the composite defect lies at the intersection of evenness, symmetry breaking, and interval diversity.
    }
    \label{fig:umap_defect_manifold}
\end{figure}

The UMAP embedding in Fig.~\ref{fig:umap_defect_manifold}A--C shows that the full root-present 12-TET scale space does not form a featureless cloud, but decomposes into discrete ``islands'' of closely related scales.  Because UMAP is applied directly to the binary pitch-class masks, points are neighbors precisely when their on/off patterns differ by only a few notes.  Coloring by evenness defect (Fig.~\ref{fig:umap_defect_manifold}A) reveals that each island corresponds to a combinatorial family (e.g.\ primarily $k\approx 5$, $6$, $7$, or $8$ notes), within which evenness varies smoothly in a mid-range ($\sim 0.4$--$0.7$).  Evenness therefore behaves as a \emph{local} order parameter on these families, rather than defining a single global axis across the manifold.  In contrast, normalized interval-entropy (Fig.~\ref{fig:umap_defect_manifold}B) is essentially saturated for most scales: almost all islands are high-entropy, with only a small number of highly regular or extremely sparse patterns at the periphery showing low entropy.  Entropy alone thus fails to stratify the manifold and motivates the use of richer geometric defect measures.

To understand how these islands relate to scale size and defect, we clustered the UMAP coordinates with $k$-means ($k=16$; Fig.~\ref{fig:umap_defect_manifold}C) and computed summary statistics for each cluster.  The clusters fall into clear bands in scale size $k$ (e.g.\ modes at $k\approx 4$--$5$, $5$--$6$, $6$--$7$, $7$--$8$, $8$--$9$ notes), and within each band they differ primarily in evenness and composite defect.  For example, one 6--7-note cluster has relatively low evenness and composite defects ($\overline{e}\approx 0.41$, $\overline{c}\approx 0.29$) and high normalized entropy ($\overline{H}_\mathrm{norm}\approx 0.90$), corresponding to comparatively regular yet information-rich scales.  Another cluster, dominated by 8--9-note scales, exhibits the highest evenness and composite defects ($\overline{e}\approx 0.72$, $\overline{c}\approx 0.43$) and markedly lower entropy ($\overline{H}_\mathrm{norm}\approx 0.69$), corresponding to highly fragmented, irregular patterns.  A third cluster containing mostly 4--5-note scales still shows relatively high composite defect ($\overline{c}\approx 0.45$), illustrating that strong geometric imperfection can arise even in very small pitch-class sets.  Remarkably, the mean Zeitler defect (number of missing fifths) is close to three in all clusters, indicating that essentially the entire manifold lives in a mid-functional-defect ``selective imperfection'' corridor; the clusters differ not in being ``too perfect'' or ``too broken'' tonally, but in how their geometric imperfections are organized.

The correlation structure in Fig.~\ref{fig:umap_defect_manifold}D clarifies these relationships.  Scale size shows only modest correlation with geometric defects, and Zeitler defect is only weakly correlated with evenness, arrangement, or interval diversity, confirming that it captures a largely independent tonal notion of imperfection.  Composite defect, by construction, lies at the intersection of evenness, symmetry breaking, and unique-intervals defects, and is strongly correlated with sequence complexity (LZ) and interval diversity but only moderately with entropy.  Together, these analyses show that UMAP organizes the scale space into combinatorial families indexed by $k$ and pitch-class content, while our geometric defect measures provide complementary, structure-sensitive coordinates within this manifold, precisely in the mid-defect regime where musically useful scales tend to reside.

\begin{figure}[ht]
\centering
\includegraphics[width=0.7\textwidth]{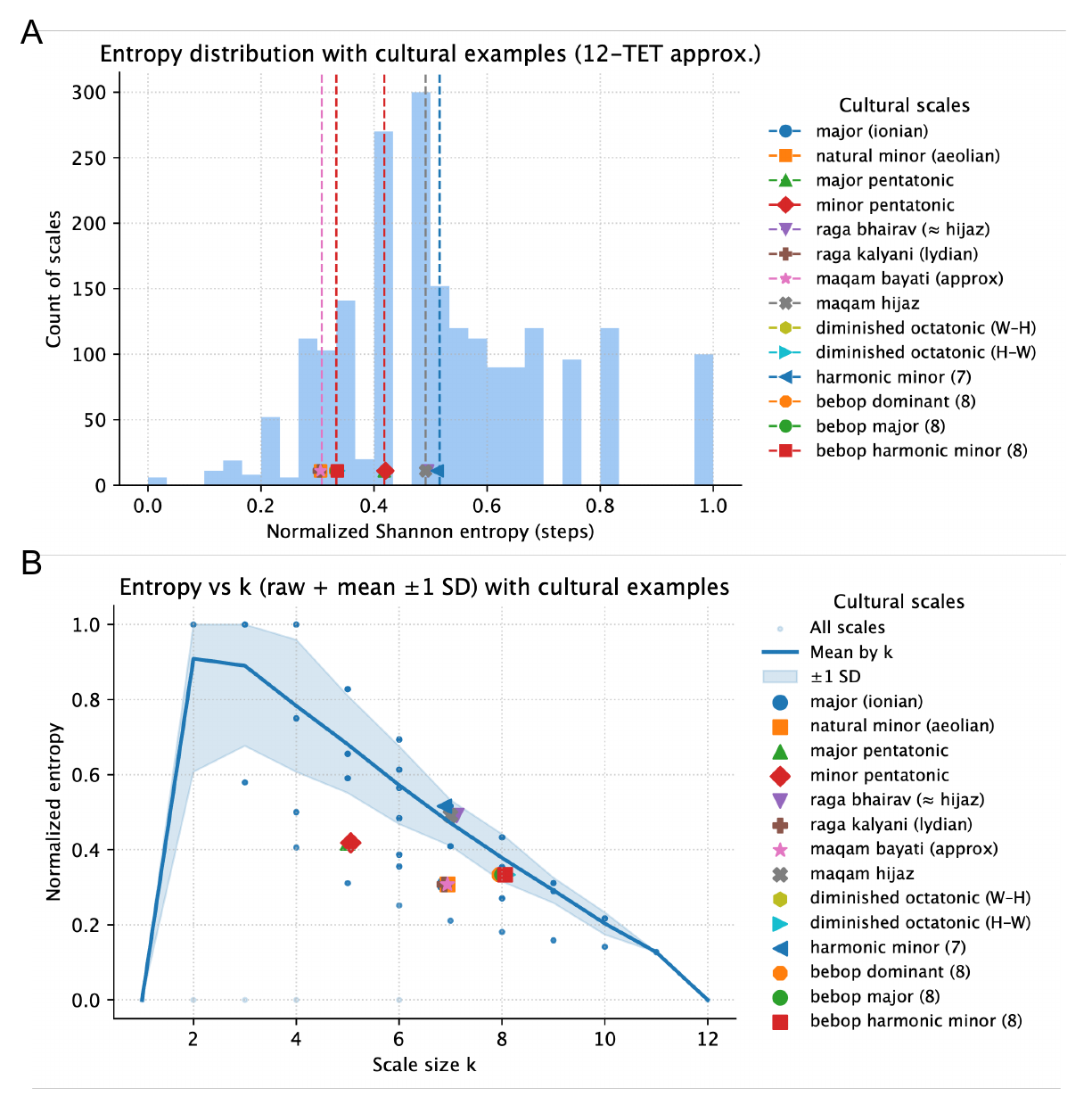}
\caption{
Cultural scales within the entropy landscape. 
(A) Histogram of normalized Shannon entropy across all valid 12-TET scales, with culturally 
salient examples indicated by colored markers and dashed lines. 
(B) Entropy versus scale size $k$, with the same cultural scales highlighted against the 
background of all possible scales. Despite the vast combinatorial space, widely used 
systems (including major and minor pentatonics, the diatonic modes, harmonic minor, 
bebop extensions, octatonic scales, and selected raga and maqam families) consistently 
fall within an intermediate entropy regime. This clustering supports the principle of 
selective imperfection introduced above, showing that both cultural evolution and theoretical design 
gravitate toward scales that balance order and surprise. 
}
\label{fig:entropy_cultural_combo}
\end{figure}

The universality of this principle is underscored by cultural practice. As shown in Fig.~\ref{fig:entropy_cultural_combo}, canonical scales from Western classical, jazz, and non-Western traditions consistently cluster within the mid-entropy regime, even though they span different sizes ($k=5$–8) and historical origins. The major and minor pentatonics, diatonic modes, harmonic minor, bebop scales, octatonics, and selected raga/maqam systems all inhabit the same corridor of moderate entropy that dominates the mathematical distribution. This convergence indicates that cultural evolution, like material self-organization, selects scales that exploit the generative ``sweet spot'' between predictability and incoherence, reinforcing the broader claim that selective imperfection is a universal driver of richness.

This clustering mirrors physical principles in materials science. The distribution of entropy across scales resembles a density of states, where most possibilities cluster around intermediate disorder, and only a few occupy the extremes. Cultural scales appear as ``stable phases'' in this landscape, analogous to the way matter self-organizes into crystalline or amorphous forms. Just as the Hall–Petch relation shows that intermediate defect densities maximize resilience, the entropy analysis shows that intermediate irregularity maximizes cultural viability. In both domains, selective imperfection transforms constraint into generativity: proteins fold, materials toughen, and scales endure not by avoiding entropy but by harnessing it in balance with order.

\begin{figure}[h!]
\centering
\includegraphics[width=0.85\textwidth]{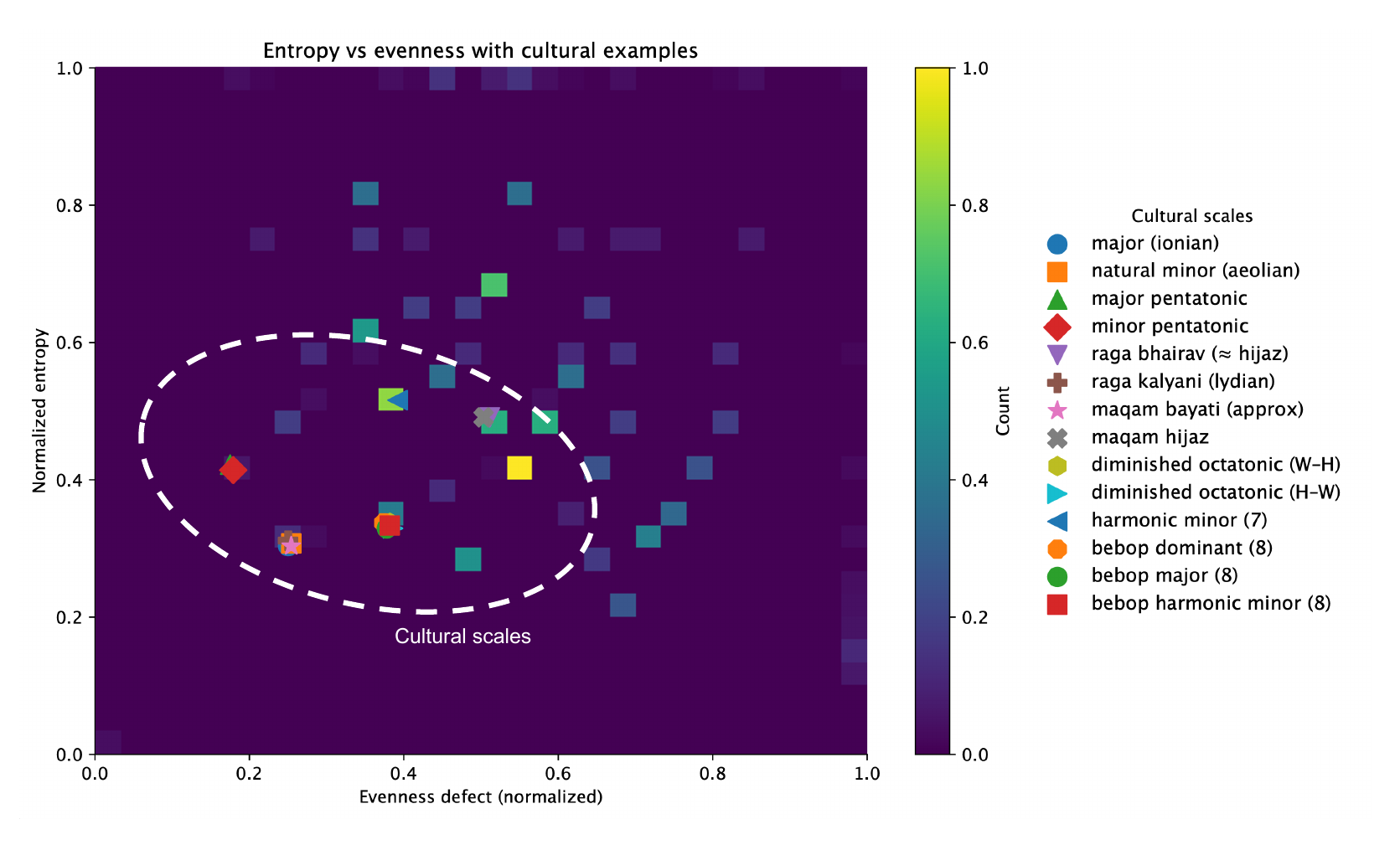}
\caption{Entropy versus evenness defect across the space of valid 12-TET scales, with cultural exemplars overlaid. 
The dashed ellipse highlights the ``cultural corridor'' of moderate entropy and defect, where nearly all salient systems 
(major, minor, pentatonic, raga, maqam, bebop) cluster. This region mirrors the material optimum of Fig.~\ref{fig:defects}, 
where intermediate defect densities maximize resilience, and connects directly to the entropy distributions in 
Fig.~\ref{fig:entropy_cultural_combo}. This provides a direct quantitative link between defect-mediated resilience in matter and imperfection-driven expressivity in music. The result leads also towards a unified view in which both material toughness and musical expressivity 
arise most robustly at intermediate levels of imperfection.}
\label{fig:entropy_evenness_bridge}
\end{figure}

Building directly on the scale enumeration of Fig.~\ref{fig:defects} and the entropy distributions of 
Fig.~\ref{fig:entropy_cultural_combo}, we plot a joint entropy–defect landscape for the musical scales (Fig.~\ref{fig:entropy_evenness_bridge}). 
This analysis unifies the two perspectives, showing that scales of moderate irregularity also display intermediate entropies, 
and it is precisely in this corridor that cultural exemplars consistently appear. 
The result parallels the Hall–Petch optimum in materials science: resilience in matter and expressivity in music both emerge 
most robustly at intermediate levels of imperfection. 
This bridge reinforces the universality of selective imperfection as a generative design principle across domains.

The analogy between defects in matter and broken symmetries in music underscores a generative principle: richness arises from selective imperfection. In materials, defects create pathways for energy dissipation and adaptation; in music, they generate tension, emotional color, and narrative flow. Both domains demonstrate that perfection alone is sterile, while imperfection, artfully deployed, becomes a source of creativity, resilience, and novelty. In this sense, defects are not aberrations to be eliminated but essential design elements that expand the expressive capacity of both matter and music.
In other words, selective imperfection functions as a generative algorithm, a mechanism through which systems escape equilibrium. In this view, entropy and asymmetry act as engines of creativity: by perturbing order just enough to provoke reorganization, they enable systems, whether metallic lattices or musical scales, to explore new regions of possibility space.

This principle extends beyond matter and music and perhaps, to the logic of cognition itself. In formal systems, as Gödel showed, completeness and consistency cannot coexist~\cite{Godel1931,Lakoff1999,Rosen1985}; only by stepping outside a closed set of rules can new truths be generated. Likewise, creative systems (whether neural, cultural, or material) derive their vitality from operating at this boundary between order and breakdown. Selective imperfection thus defines a universal generative mechanism: it breaks symmetry just enough to invite emergence, allowing structures, melodies, and ideas to reorganize into higher forms of coherence. What appears as defect or deviation is therefore not noise but the seed of transformation, the very point where the another world begins to compose anew.

\begin{figure}[h!]
\centering
\includegraphics[width=0.65\textwidth]{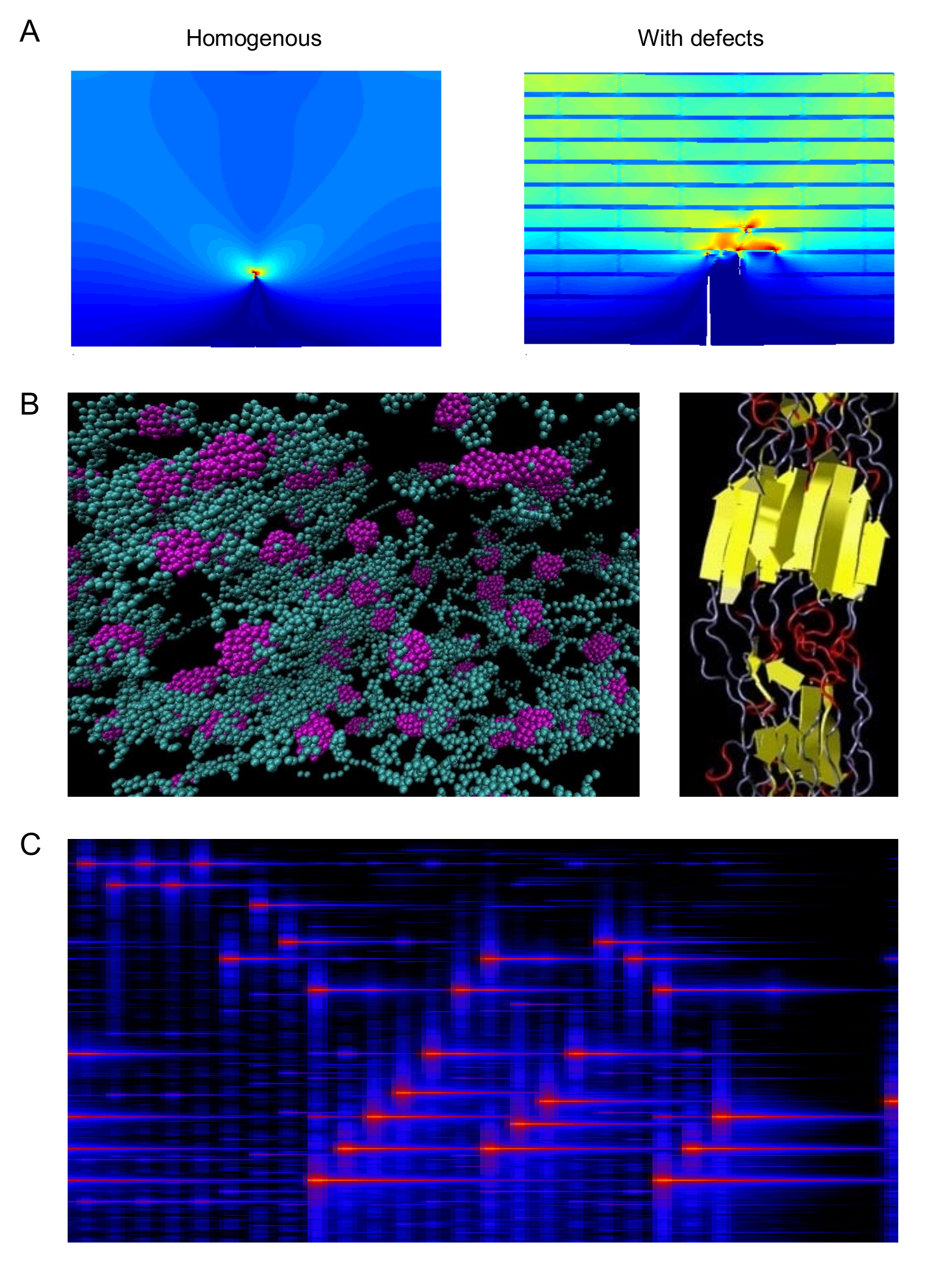}
\caption{Higher-order compositional heterogeneous structures in matter and music.
(A) Stress distribution in a composite without defects (left) versus with
distributed defects (right)~\cite{Dimas2013AdvMater}. The homogeneous case shows strong localization
of stress near the crack tip, leading to catastrophic fracture, whereas
defects deflect and redistribute stress, enhancing toughness and resilience.
(B) Atomistic rendering of spider silk microstructure, highlighting the
heterogeneous organization of crystalline (magenta) and amorphous (cyan)
domains that together provide a balance of strength and extensibility. A molecular-level nanostructure of silk proteins is shown on the right, revealing similar heterogeneous structuring~\cite{Keten2010AtomisticNanostructure}. 
(C) Melodic spectrogram of Beethoven's \emph{Für Elise}, showing the heterogeneous
melodic landscape: recurring motifs, tonal clusters, and variations in pitch
distribution that break uniformity and create musical richness. Together,
these examples illustrate how heterogeneity and controlled imperfection
disperse concentration, prevent failure, and generate expressive depth in
both materials and music.}
\label{fig:heterogeneous}
\end{figure}

Figure~\ref{fig:heterogeneous} highlights how heterogeneity itself becomes a design principle. In materials, distributed defects in a composite disperse stress and deflect cracks, preventing catastrophic failure and enhancing toughness (panel A), while the nanoscale arrangement of crystalline and amorphous domains in silk provides a balance of strength and extensibility (panel B). Music exhibits a parallel form of heterogeneity: the spectrogram of Beethoven’s \emph{Für Elise} (panel C) reveals clustered motifs, tonal variations, and departures from uniformity that generate structure and expression. In both domains, richness and resilience emerge not from homogeneity, but from carefully organized heterogeneity.

\begin{figure}[ht]
\centering
\includegraphics[width=0.85\textwidth]{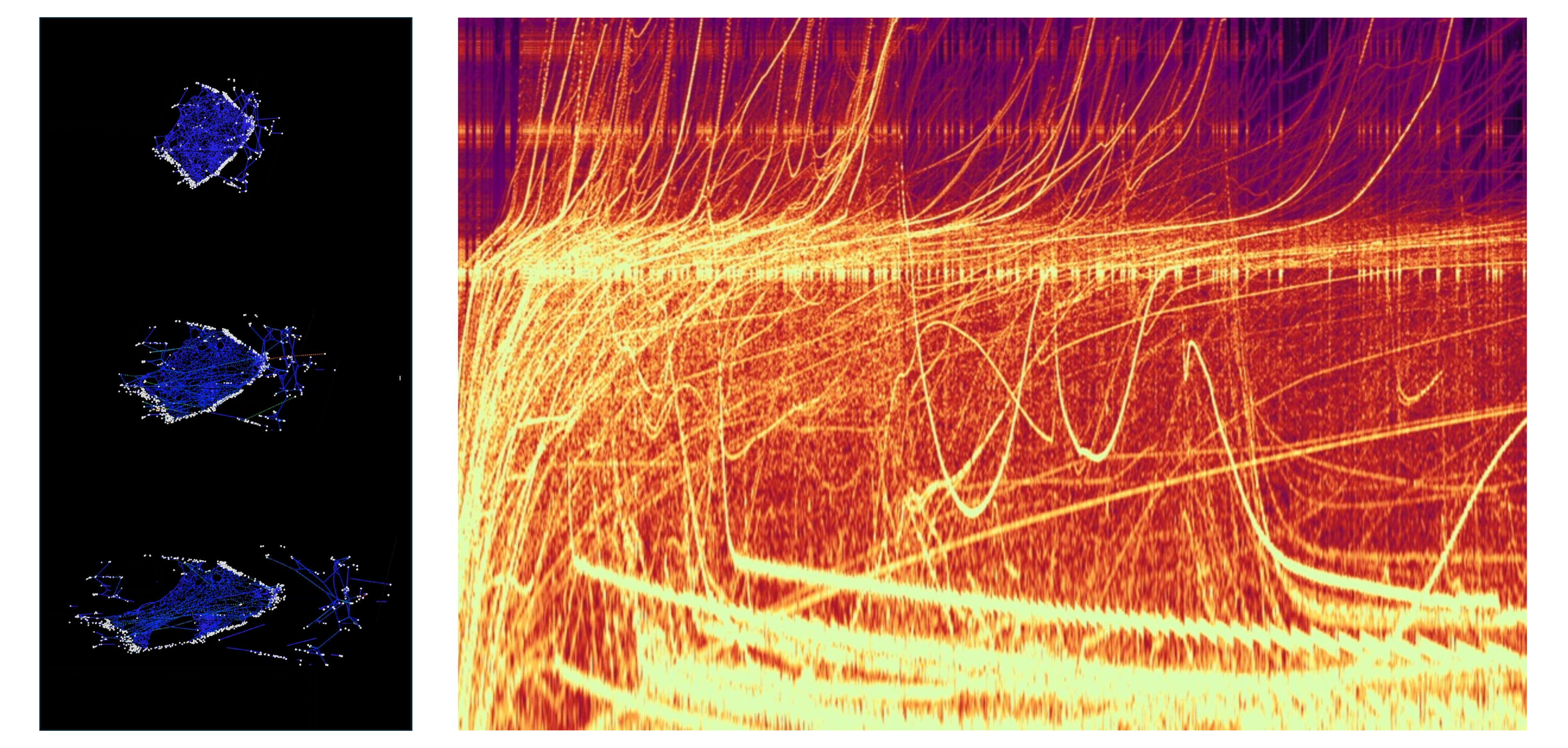}
\caption{Spider web sonification under stress (left, visualization of stretching, right, sonic spectrum evolution during stretching)~\cite{Su2018JRSi,Su2021CMJ,Su2022JMUI}.
Spectrogram representation of a spider web as it is loaded mechanically. 
As stress increases, vibrational frequencies rise (upward-moving streaks), 
reflecting tension-induced stiffening of silk fibers. When fibers break, 
frequencies drop or vanish (abrupt terminations of tracks), producing 
distinct spectral discontinuities. The web thus ``sings'' its own structural 
state, with stress and fracture rendered as audible changes in frequency, 
illustrating how spider silk embodies both material resilience and musical 
expression. These mappings establish a form of cross-species epistemology, where vibration functions as a shared physical syntax enabling dialogue between distinct perceptual worlds. This framework reframes sonification as translation between ontologically diverse but structurally homologous cognitive systems.}
\label{fig:websonification}
\end{figure}

This principle becomes especially vivid when the spider web itself is treated as an instrument. By mapping the web’s vibrations into sound, we can follow how its acoustic landscape evolves under load: frequencies rise as tension stiffens the fibers, and they collapse when strands rupture (Fig.~\ref{fig:websonification}). The web thus ``sings'' its structural state, transforming stress and fracture into audible events that mirror the generative role of imperfection in both matter and music.

Naturally, this principle extends beyond materials and music: selective imperfection represents a general algorithmic mechanism through which complex systems escape equilibrium to generate novelty.

\subsection{Isomorphisms and Sonification}

A central principle of materiomusic is the recognition of isomorphisms between structures in matter and forms in music. In mathematics, an isomorphism denotes a structure-preserving mapping between two domains; here, it describes how molecular hierarchies or architectural webs can be translated into musical systems without loss of relational information. Proteins provide one example: amino acid sequences map to pitch classes through a self-consistent vibrational scale, preserving their sequential and modular architecture while making them audibly perceptible \cite{Yu2019ACSNano,Qin2019EML}. This mapping is reversible, enabling not only the sonification of proteins but also the compositional design of \textit{de novo} sequences from musical inputs, transforming motifs and harmonic rules into foldable molecular candidates~\cite{DeepAria2021}.  A practical key to this bridge is transpositional equivalence: by shifting vibrational information across frequency bands while preserving ratios, patterns remain recognizable to perception. This enables molecular vibrations far beyond human hearing to be rendered as stable musical objects, maintaining relational structure while crossing otherwise incommensurate time and frequency scales.  

Listening in this framework is not merely representational but analytical. 
To sonify a protein or a spider web is to construct an auditory model that 
preserves functional relationships, such that differences in pitch, rhythm, 
or timbre correspond to differences in structure or function. In this sense, 
listening becomes a form of seeing: an epistemic inversion where auditory 
perception becomes a scientific instrument for probing categories and 
hierarchies otherwise hidden from view.

Spider webs extend this concept into three-dimensional architecture. Each silk fiber can be modeled as a vibrating string whose pitch is determined by its tension, length, and diameter. When scanned and reconstructed as networks, webs can be transposed into instruments in which fibers become musical voices, allowing the geometry of a web to be performed and explored sonically~\cite{Su2018JRSi,Su2021CMJ,Su2022JMUI}. As threads are stretched or broken, their vibrational frequencies shift, making mechanical change audible in real time. The isomorphism here preserves hierarchy (nodes, threads, modules, and the global web) as layers of rhythm, harmony, and form, rendering the web’s structure both a score and an instrument.  This inversion also highlights listening as a mode of seeing. This is another example for sonification that functions not merely as representation but as epistemic inversion so that auditory perception becomes a scientific instrument for probing categories and hierarchies that remain opaque to visual or statistical analysis. In this framework, to listen to a protein, web, or flame is to conduct an experiment with the ear, revealing residues of structure and deep-time constraints that conventional methods might overlook.

Perception itself is a translation. The spider’s leg hairs sense frequency bands far below human hearing; our ears, in turn, compress wide vibrational spectra into a narrow auditory window. Each system filters the same physics through species-specific constraints. By sonifying spider webs (or other biologically generated systems), we construct an intentional translation, an interpretive bridge, mapping an invertebrate’s vibrational universe into the human sensory domain. The aim is not to imitate the spider’s perception, but to render its world legible to ours, creating a dialogue between species through the medium of vibration.

These sonifications illustrate the broader potential of isomorphisms: they provide a reversible grammar for exploring matter and music as dual instantiations of the same generative principles. By moving across domains while preserving relational structure, they allow scientific analysis, artistic performance, and material design to converge into a single language of vibration.

The broader historical trajectory of computer music also informs this work. Early analog synthesis in the mid–20th century emphasized direct manipulation of waveforms through oscillators, filters, and voltage control, allowing composers to sculpt timbre from first principles of physics. Instruments such as the RCA Mark II or later modular systems by Moog and Buchla embodied a generative paradigm, in which sound was built bottom-up from continuous signals rather than discrete symbolic notation. The subsequent rise of digital synthesis and sampling in the 1970s–1980s shifted the emphasis toward algorithmic control and memory, culminating in the personal computer and software studios that made electronic composition widely accessible. More recently, granular synthesis, physical modeling, and machine-learning-based systems have opened new vistas for composition, extending the analog ethos of direct sound manipulation into domains where models of instruments, materials, or even proteins can be rendered audible. This historical arc situates materiomusic as a continuation of computer music’s evolution: a translation of molecular and structural data into sound that parallels the trajectory from analog waveforms to digital computation and now to agentic AI.

\begin{figure}[ht]
\centering
\includegraphics[width=0.8\textwidth]{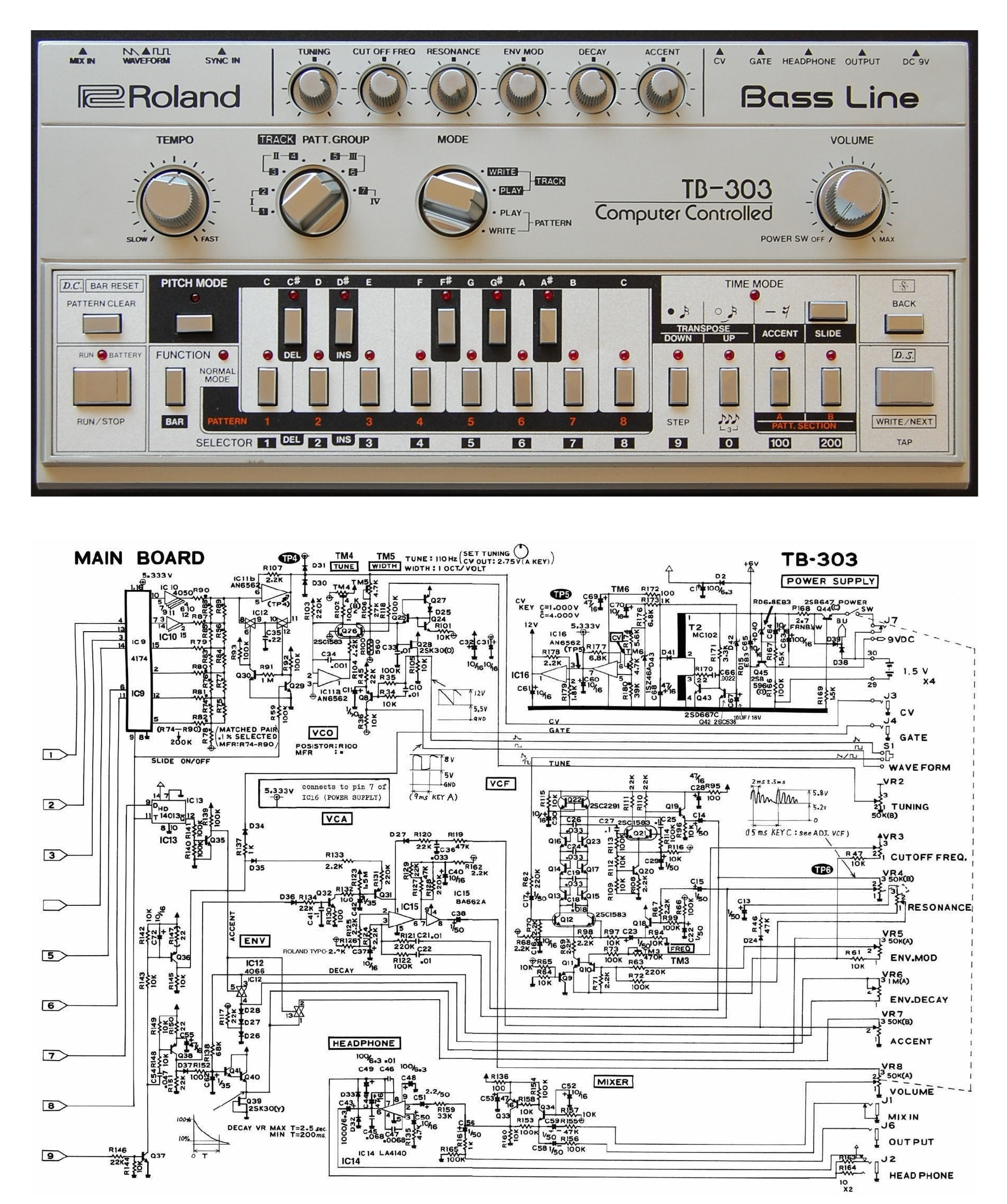}
\caption{The Roland TB-303 as a generative instrument, showing how generative principles can be encoded directly in matter, here through the nonlinear interplay of oscillators, filters, and feedback in circuitry (reprinted from~\url{https://synthfool.com/docs/Roland/TB303/}).
(A) Front panel of the TB-303 synthesizer (Roland, 1981), whose limited 
controls for tuning, cutoff frequency, resonance, envelope modulation, 
decay, accent, and step programming enabled the creation of iconic 
bassline patterns. 
(B) Main board schematic showing the internal generative architecture: 
a voltage-controlled oscillator (VCO), filter (VCF), amplifier (VCA), 
and envelope (ENV), coupled through accent and slide circuits and 
feedback loops. 
The TB-303 illustrates how material implementations of oscillators, 
filters, and nonlinear feedback serve as generative algorithms for sound. 
Its quirks and constraints, once seen as defects, became creative 
affordances that shaped entire musical genres.}
\label{fig:tb303}
\end{figure}

A particularly emblematic example is the Roland TB-303, introduced in 1981 as an analog bass synthesizer with built-in computer sequencing. Originally intended as an automated substitute for bass guitar, its squelching resonant filter and step-programmed patterns were initially considered a commercial failure. Yet in the late 1980s, artists such as Phuture in Chicago reappropriated the TB-303, exploiting its imperfections and nonlinear resonance to define the sound of acid house. Here, the interplay of analog circuitry and algorithmic sequencing produced emergent aesthetics far beyond the designers’ intent. This trajectory, where an instrument conceived as imitation becoming a generator of new musical genres, mirrors the generative materiomusic ethos: mappings from structure to sound that are not merely representational but transformational, yielding novelty from constrained rules and system imperfections that force creators to expand the rules of the game~(Fig.~\ref{fig:tb303}).

The Roland TB-303 also illustrates how generative principles can be encoded directly in ``analog'' matter, here through the nonlinear interplay of oscillators, filters, and feedback in circuitry. What appears as a set of simple knobs and sequencer steps (Fig.~\ref{fig:tb303}A) is underpinned by a network of material algorithms (Fig.~\ref{fig:tb303}B) that produce emergent sonic behaviors. Just as spider webs or protein vibrations translate physical oscillations into musical structure, the TB-303 demonstrates how designed heterogeneity and even imperfections become creative affordances, reinforcing materiomusic as a design language.

It is possible to implement composition via stigmergic swarms: agents leave traces in a
shared field (digital pheromones / musical motifs) that others sense and adapt
to, akin to an improvising ensemble with indirect coordination. Roles emerge,
redundancies fade, and long-range coherence develops. Network analysis of the
outputs shows human-like signatures: high small-worldness, balanced modularity,
and cross-community participation, indicating structure that is locally varied
yet globally integrated, notably beyond surface interpolation.

These analyses are possible because at its foundation, the convergence of matter and music can be expressed through mathematics. Both are governed by discrete alphabets (twenty amino acids, twelve pitch classes) and by generative rules that yield hierarchical, multiscale structures. These alphabets can be formalized in terms of group theory and modular arithmetic, where transposition, inversion, and permutation act on residues or tones, while folding pathways and counterpoint progressions are isomorphic to graph-theoretic trajectories across energy landscapes. The statistical universals observed in protein families and musical scales alike can be described through information theory, where entropy balances redundancy and surprise, and through network theory, where measures such as small-worldness, modularity, and degree distribution quantify structural coherence. More fundamentally, category theory provides a unifying abstraction: amino acid sequences and musical motifs are morphisms in compositional categories, with functors mapping vibrational data to auditory spectra, or graph embeddings to harmonic relations. This rigorous formalism shows that materiomusic is not a metaphor but a mathematically consistent language, where structure-preserving mappings translate deep-time information across physical, biological, and cultural domains.

Our reversible map enables a composition-as-design workflow: define a vibrationally
consistent scale for the 20 amino acids; compose musical material that enforces
long-range relations (motif reuse, thematic return); invert to sequences; and
screen candidates via structure prediction prior to synthesis and characterization.
Crucially, this allows targeted perturbations, such as inserting a small ``imperfection''
in the melody, in order to serve as testable mutations that can rebalance stability or
function.

\section{AI and the Question of Creativity}
\subsection{Limits of Conventional AI}
\begin{figure}[ht]
\centering
\includegraphics[width=0.85\textwidth]{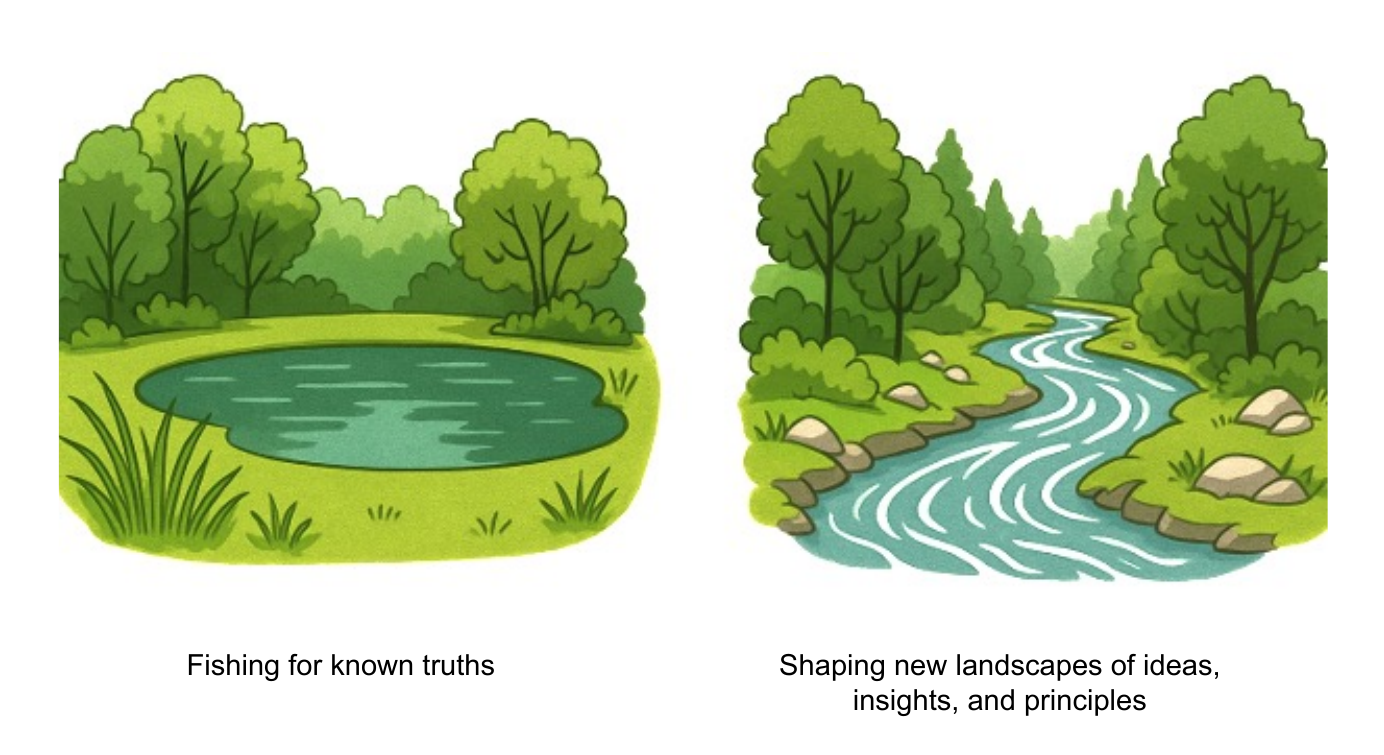}
\caption{%
Two modes of discovery. Conventional approaches, whether in science or AI, often resemble 
fishing in a fixed pond of known truths; these methods are efficient at retrieval but limited to what already exists. 
In contrast, generative frameworks such as materiomusic and agentic swarm AI aim to shape new landscapes 
of ideas, insights, and principles, extending beyond interpolation into invention that shape the world itself. 
This shift underpins the broader claim that discovery is not only about accessing knowledge, 
but also about composing new structures within the open-ended grammar of matter and music.
}
\label{fig:fishing_landscapes}
\end{figure}

Most contemporary AI systems excel at interpolation: they generate outputs that lie within the statistical envelope of their training data. This makes them powerful tools for prediction, pattern recognition, and replication of known styles, but it also constrains them to reconfigurations of existing truths rather than the invention of genuinely new ideas. In the sciences, this means that large language models or generative systems can summarize known principles and propose plausible extensions, yet they struggle to originate hypotheses that break free of precedent or challenge established beliefs. In music, they can emulate styles or stitch together patterns, but their outputs often lack the long-range coherence, closure, and surprise that characterize human creativity (Fig.~\ref{fig:fishing_landscapes}).  

The mathematician Kurt Gödel demonstrated that in any sufficiently complex formal system, there exist true statements that cannot be derived from the rules of the system itself. By analogy, AI models trained on finite datasets embody incomplete formal systems: they can recombine what they have seen, but they cannot, without augmentation, transcend the space defined by their training distribution. As a result, they act more like mirrors of past knowledge than engines of invention. To move beyond interpolation requires a shift in paradigm from static models that optimize likelihood, toward dynamic, self-organizing systems that thrive in the open-ended space of unsolved problems by expanding the set of axioms that define their world~\cite{Buehler2025MusicSwarm}.

Gödel’s insight may parallel the role of intuition in human creativity. Akin to the fact that no formal system can prove all truths within its axioms, no fixed model can foresee all patterns that will emerge in the open world. Human intuition (often experienced as aesthetic judgment or sudden insight) serves as an informal oracle, bridging gaps that logic cannot. Embedding such open-endedness in AI requires architectures that learn to sense rather than to solve, to listen for resonance rather than to optimize. Agentic swarms emulate this process by evolving intuition collectively through interaction, rather than encoding it explicitly.

Recognizing these limits is essential for reframing AI not as a parrot of human knowledge but as a partner in discovery. By incorporating principles of generativity, imperfection, and collective dynamics, we may design agentic AI systems capable of venturing into the incomplete, unexplored domains where novelty arises. This is also why conventional, monolithic AI struggles with deep time: it lacks mechanisms to reconcile femtosecond molecular patterns with human-scale cultural forms. To traverse such spans, AI must operate as a translator across scales) maintaining relational structure while moving between modalities and temporal regimes (rather than as a static interpolator within a single distribution. 

It is also essential to emphasize that while AI provides unprecedented tools for traversing scales and interrogating hidden structures, it does not replace the human creative act. Rather, AI functions as a collaborator - translating inaccessible molecular vibrations or evolutionary residues into audible structures - while humans provide interpretation, aesthetic judgment, and contextual meaning. This partnership ensures that deep-time information, whether emerging from proteins, webs, or fractures, is not only rendered audible but also integrated into cultural and scientific creation.  

It is important to emphasize the division of labor between humans and agentic AI. Humans provide framing, intention, aesthetic judgment, and ethical context, while AI acts as a collaborator that translates across scales and modalities. In this partnership, AI surfaces patterns hidden in molecular vibrations or web topologies, but humans supply interpretation and cultural meaning. This balance ensures that materiomusic remains both scientifically rigorous and artistically resonant.

\subsection{Collective Intelligence Models}

One path beyond the limits of conventional AI lies in embracing collective intelligence~\cite{ghafarollahi2025sparksmultiagentartificialintelligence,Buehler2025MusicSwarm}. Rather than relying on a single, monolithic model, swarm-based frameworks deploy ensembles of agents that begin from shared rules but differentiate dynamically through interaction. In this setup, each agent acts as a contributor to a shared composition, guided by local objectives, memory, and feedback, while global structure emerges from their interplay. This mirrors processes in biology, where collective behavior (such as ant colonies, bird flocks, or the assembly of proteins) generates complexity without central control.

In the Sparks model~\cite{ghafarollahi2025sparksmultiagentartificialintelligence}, for instance, the authors operationalize the thermodynamic definition of creativity defined in this paper through an open-ended exploration engine that continuously constructs and stress-tests a scientific world model. By deploying adversarial agents to intentionally ``break'' the current model's constraints, the system forces the emergence of new theoretical frameworks. This is akin to introducing new degrees of freedom that are necessary to re-establish stability in an expanded solution space. The agents build a world model, encounter adversarially induced constraint failures, and are mechanically forced to invent new principles. This adversarial dynamic propels the AI beyond simple interpolation, compelling it to invent genuinely new principles, such as the beta-sheet crossover in proteins as discussed in~\cite{ghafarollahi2025sparksmultiagentartificialintelligence}, as a survival response to induced conceptual failure. Related efforts used transitive and isomorphic graph operations~\cite{Buehler2024Accelerating,https://doi.org/10.1002/adma.202413523} to bridge disparate conceptual domains and introduce new exploratory pathways in AI systems. Therein, we replace the ``interpolative'' limitations of standard Deep Learning with ``generative inquiry'', where the graph-based world model serves as a thermodynamic gradient, providing the external perspective necessary to transcend Gödelian incompleteness.

\begin{figure}[h!]
\centering
\includegraphics[width=0.7\textwidth]{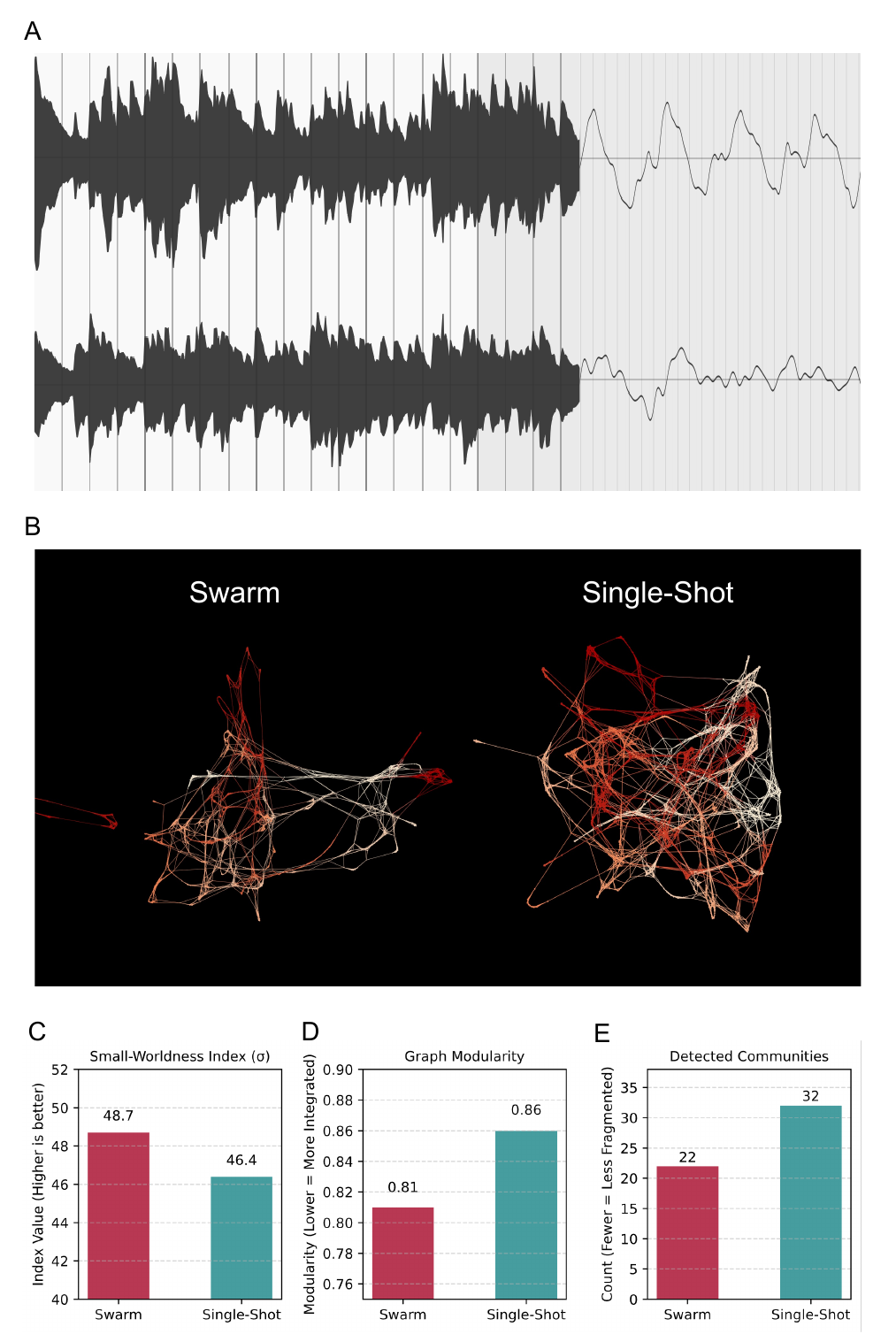}
\caption{Swarm-generated music and its network structure as reported in~\cite{Buehler2025MusicSwarm}.
(A) Waveform of the generated music (sample; left longer time-range, right zoomed in view).
(B) Network visualizations of motif connectivity derived from the same systems. The swarm produces denser and more globally integrated structures, where motifs interconnect across the network, while the baseline systems fragment into separate clusters.
(C) Quantification of small-worldness ($\sigma$), showing that the swarm achieves markedly higher values, a hallmark of complex, organic systems.
(D) Graph modularity (greedy), indicating that the swarm networks are less segregated and more globally woven together, in contrast to the higher modularity of the baselines.
(E) Count of detected communities (Louvain), where the swarm yields fewer partitions, confirming that it does not simply repeat or segregate motifs but weaves them into a connected whole. These data collectively show that the swarm yields networks resembling the locally varied, globally integrated character of human music. Evaluation details see original paper~\cite{Buehler2025MusicSwarm}.}
\label{fig:swarm_music}
\end{figure}

Related recent work on MusicSwarm illustrates the power of this approach when scaled to massively agentic systems, and addressing specifically creative tasks such as composition. Agents in the swarm communicate indirectly through a shared ``pheromone field,'' a stigmergic medium in which they leave traces of motifs, structures, or ideas. Other agents respond to these signals by reinforcing, mutating, or extending them, creating feedback loops that drive differentiation and emergent specialization \cite{Buehler2025MusicSwarm}. Some agents evolve toward long-range structural integration, while others focus on local motifs or rhythmic variations. The swarm therefore balances diversity with coherence, producing outputs that resemble the long-range memory, closure, and thematic development typical of human composition (Fig.~\ref{fig:swarm_music}).  

A crucial feature of collective intelligence is adaptability. Because each agent continuously updates its traits in relation to others, the swarm converges toward equilibria reminiscent of game-theoretic dynamics, where no agent has the incentive to deviate. This self-organization fosters resilience: if one agent fails or collapses into repetition, others can compensate, preserving global creativity. More importantly, novelty emerges not from pre-training on vast datasets but from the dynamic exchange of partial solutions and their recursive recombination. These ideas parallel the principles laid out in \textit{Agentic Deep Graph Reasoning}, where recursive updates in knowledge networks produce scale-free, small-world topologies and reveal the incompleteness of any single formal representation \cite{Buehler2025ADGR}. Both in graph reasoning and in music swarms, invention arises from distributed dynamics that weave together local constraints and global coherence, pointing to a broader paradigm of agentic, generative intelligence.

\subsection{Evidence of Novelty}

The claim that swarm-based intelligence moves beyond interpolation toward invention can be quantitatively assessed using graph-theoretic measures applied to musical outputs. By transforming self-similarity matrices into networks, where nodes represent segments of music and edges denote thematic recurrence, we can directly compare the structural properties of human, baseline AI, and swarm-generated compositions. This analysis reveals that swarm outputs consistently exhibit signatures associated with human-like creativity.  

One key metric is small-worldness ($\sigma$), which captures the balance between local clustering and global integration. Human music typically displays high $\sigma$, reflecting coherent motifs embedded within long-range thematic arcs. Swarm compositions achieve the highest small-worldness among all AI baselines, weaving motifs into globally connected wholes rather than fragmenting into isolated modules \cite{Buehler2025MusicSwarm}. Modularity analysis further shows that while single-shot and traditional multi-agent systems often segregate into rigid communities, swarms favor cross-connections that sustain thematic flow across sections. This integration produces cadences, returns, and narrative closure, qualities largely absent from monolithic AI outputs.  

Additional metrics reinforce this picture. Participation coefficients demonstrate that motifs in swarm compositions circulate across communities rather than remaining trapped within a single section, a hallmark of thematic development. Long-range edge fractions confirm that swarm networks contain the greatest proportion of distant connections, echoing the structural memory and delayed resolution common in human composition. Temporal analyses of connectivity show swarms progressively adapt toward equilibria with broad integration, suggesting that the collective dynamics themselves are a source of emergent form.  

Taken together, these measures constitute evidence that swarm-based systems approximate not only the surface features but also the deep structural properties of human creativity. Unlike conventional models that produce either repetition or incoherent novelty, swarms generate music with organic balance: locally varied, globally integrated, and imbued with long-range coherence. These results provide quantitative support for the hypothesis that collective, agentic intelligence can act as a creative partner, achieving invention through distributed dynamics rather than monolithic optimization.

\section{Case Studies in Materiomusic}

\subsection{Protein Music and \textit{De Novo} Design}
Protein music begins from a physically grounded premise: amino acids possess characteristic vibrational signatures that can be mapped to pitch classes, yielding a self-consistent translation between sequence space and musical space \cite{Yu2019ACSNano,Qin2019EML}. This mapping makes molecular structure audible and, crucially, reversible, melodic and harmonic organization can be used as a constructive scaffold for proposing new sequences. Building on this foundation, we developed workflows in which musical materials encode protein-relevant constraints (motif reuse, long-range recall, closure), while machine learning can assist in refining sequences toward foldability and novelty.

Proteins can be understood as ancient evolutionary memories; representing molecular artifacts that carry information from billions of years of natural history. Their amino acid sequences and folding rules encode constraints accumulated across deep time, representing the physical residues of evolution’s long experiment with matter. When sonified, these vibrational signatures make audible not just a molecular structure, but the memory of life’s origins. In this way, protein music provides a unique window into evolutionary memory, turning the deep-time archive of molecular rules into perceptible and composable sound.

\begin{figure}[ht]
\centering
\includegraphics[width=0.8\textwidth]{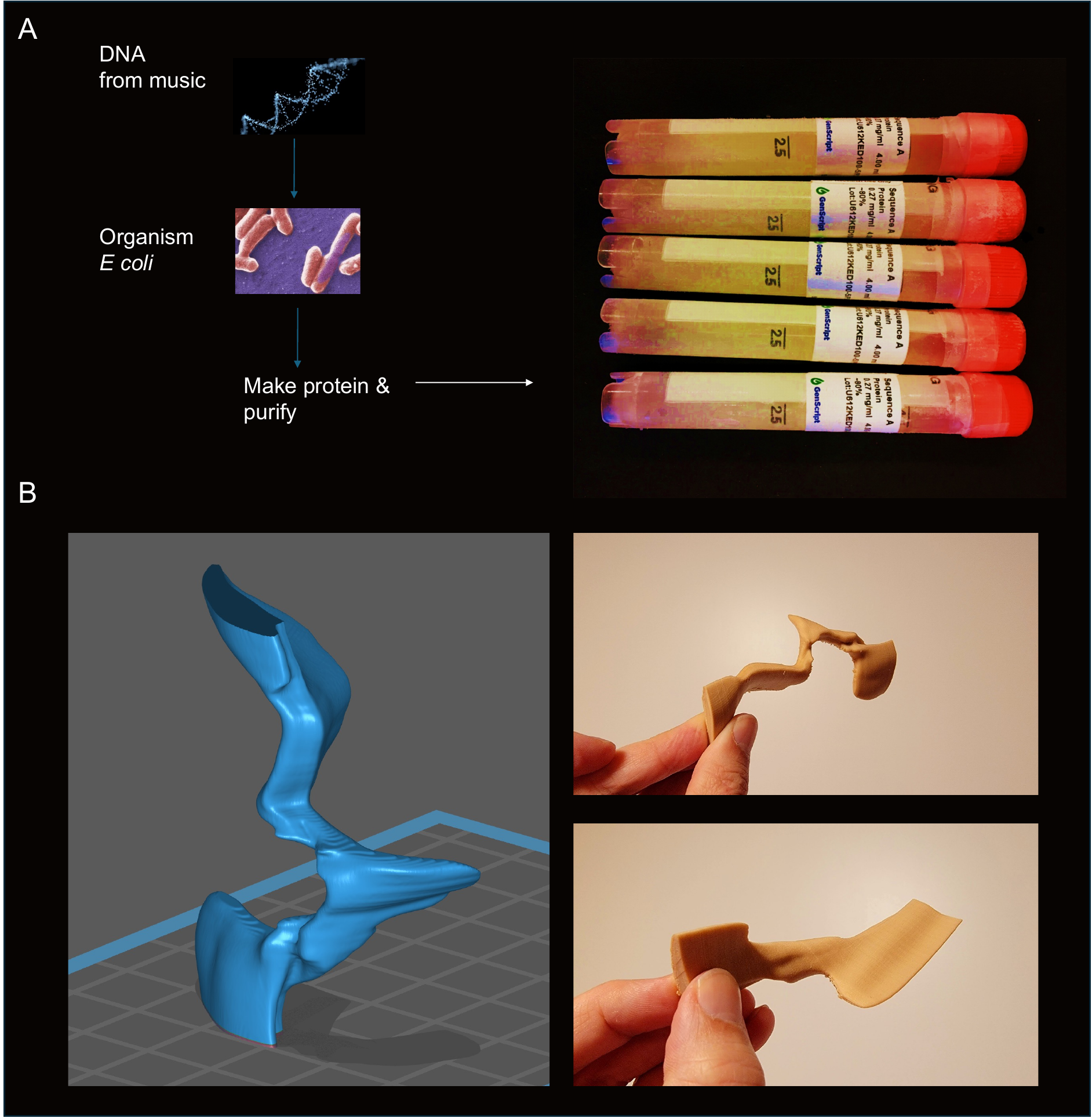}
\caption{From music to matter. 
(A) Experimental pipeline in which a musical composition is mapped to a DNA sequence, 
expressed in \textit{E. coli}, and purified as protein material. 
This workflow demonstrates the reversibility of materiomusic: melodic and harmonic structures 
become constructive templates for biomolecular sequences. 
(B) Digital rendering (left) and 3D-printed models (right) of protein folds derived from 
the same musical inputs, illustrating how compositional logic is materialized both in living 
systems and in tangible artifacts, utilizing a VAE model~\cite{Milazzo2021iScience}. Together, these steps embody the full loop from composition 
to synthesis, integrating artistic creation with experimental validation. 
}
\label{fig:music_to_matter}
\end{figure}

Deep Aria is a representative case. Starting from J. S. Bach’s Goldberg Aria, we used the amino-acid scale mapping to produce an initial \textit{de novo} sequence, then applied deep learning to elaborate variants that retain the Aria’s counterpoint-like long-range dependencies while diverging locally at the residue level \cite{DeepAria2021}. Analyses included BLAST comparisons of whole-sequence and secondary-structure fragments to verify novelty alongside structural plausibility, coupled with predictive assessments of stability and fold architecture. The result is a family of Aria-derived sequences that exhibit robust, protein-like organization while embodying the musical work’s thematic memory and hierarchical form \cite{DeepAria2021,Yu2019ACSNano}. Beyond analysis, the pipeline is experimental: sequences are reverse-translated to DNA, expressed in \textit{E.\ coli}, and purified for characterization, closing the loop from score to synthesized matter (Fig.~\ref{fig:music_to_matter}).
In the Deep Aria experiment, folding acts as a medium that injects ancient constraints into a contemporary score: the learned rules by which sequences become structure embed the evolutionary memory of proteins into the resulting music. The output is therefore not a copy of the source but a transformation that carries a residue of deep time, whereby new information added by the sequence-to-structure map that becomes audible as counterpoint, recall, and closure, and testable as a realizable \textit{de novo} fold. 

This process can be framed as a general workflow: (1) compose within a vibrationally consistent musical scale, embedding motifs and long-range recall; (2) map the composition back into amino-acid sequences; (3) evaluate candidates through structure prediction and similarity analysis; and (4) synthesize and test experimentally. Composition thus generates hypotheses, prediction provides screening, and experiment delivers validation. The entire cycle is grounded in a reversible mapping that ensures musical logic translates into molecular plausibility.

The iterative dialog between music and molecular folding operates as a concrete mechanism for traversing temporal scales. When Bach's compositional structures are mapped to amino acid sequences and subjected to folding, the process acts as an evolutionary filter: the physics of protein folding embodies billions of years of biochemical optimization, imposing constraints that Bach's original composition could not have anticipated. These constraints (such as preferences for certain secondary structures, hydrophobic core formation, loop geometries) are not arbitrary but represent solutions refined by evolution. When the folded structure is mapped back to music, it carries this evolutionary information as new compositional elements: unexpected intervallic relationships where beta-sheets form, rhythmic disruptions at loop regions, harmonic progressions influenced by disulfide bridges. This is not mystical ``memory'' but a mechanistic dialog: musical logic encounters physical law, and the collision generates genuinely new information. Each iteration through this cycle, composition to sequence to folding to sound, allows human creativity to interact with non-human constraints accumulated across geological time. The folding process thus serves as a generative oracle, transforming abstract musical relationships through the concrete grammar of molecular evolution and returning enriched compositional material that neither Bach nor evolution alone could have produced.

\begin{figure}[ht]
\centering
\includegraphics[width=1\textwidth]{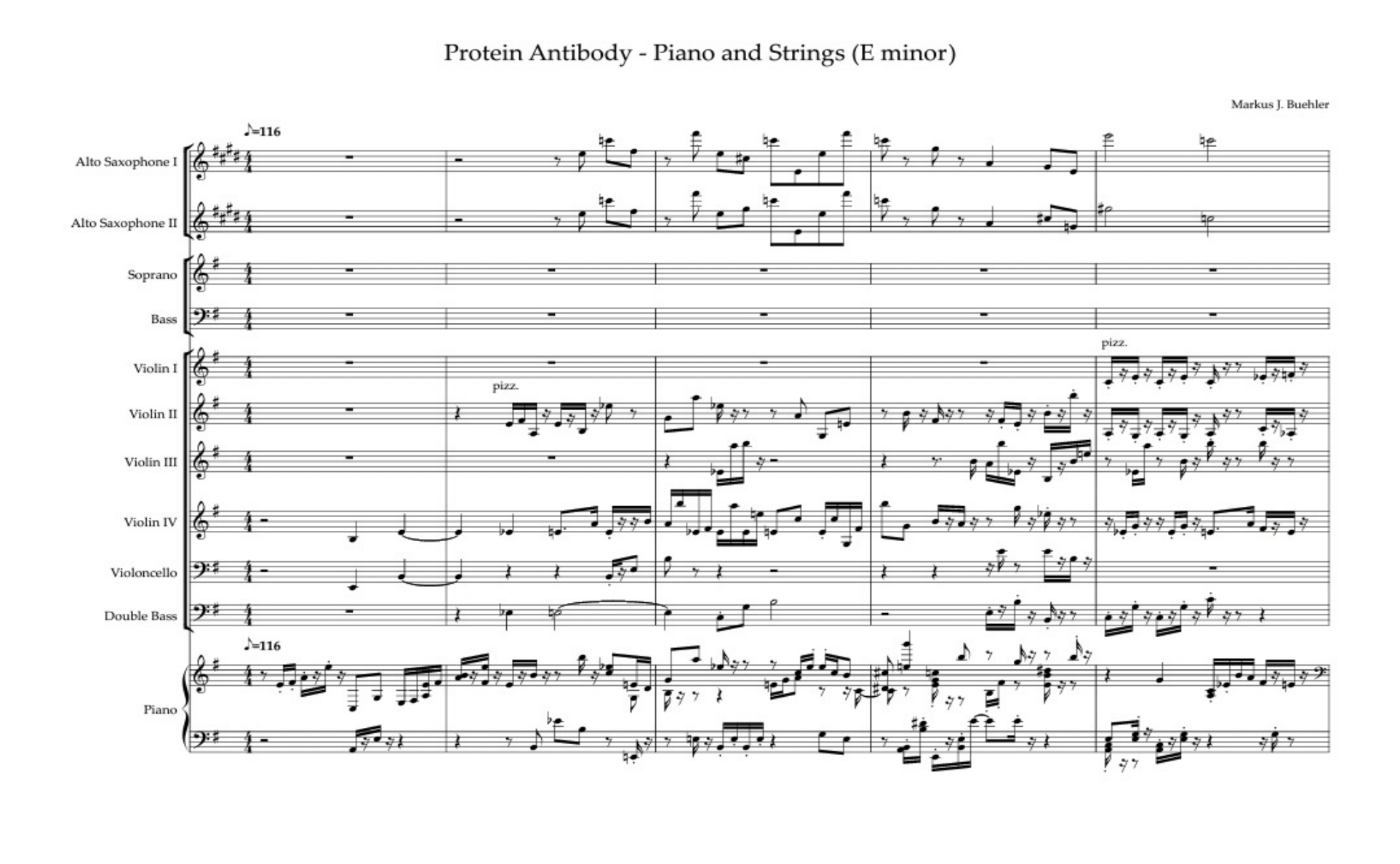}
\caption{Protein Antibody Score. Excerpt from \textit{Protein Antibody – Piano and Strings (E minor)}, 
a composition derived from immunoglobulin sequences~\cite{AntibodyMusic2020}. Here, amino acid motifs are mapped into 
orchestral textures, with strings and winds carrying different components of the construction principles. 
This example illustrates how materiomusic renders biomolecular architecture audible in a 
traditional performance medium, making hierarchical sequence constraints perceptible through 
counterpoint, orchestration, and harmonic development. This work utilized protein structure mapping as background, and constraint, for orchestral composition with artistic choices.
}
\label{fig:protein_antibody}
\end{figure}

Prior efforts illustrate the breadth of this approach. Antibody music projects mapped immunoglobulin sequences to orchestral textures to expose domain architecture and repetition through harmony and orchestration, demonstrating how sequence logic can be rendered perceptible at human temporal scales \cite{AntibodyMusic2020} (excerpt of the score, see Fig.~\ref{fig:protein_antibody}). More broadly, the sonification framework supports iterative design: sequence motives, rhythmic grouping, and polyphonic layering act as generative primitives that can be recombined, optimized with learning-based priors, and validated via structure prediction and sequence similarity screens \cite{Yu2019ACSNano,Qin2019EML}. In all cases, composition is not a metaphor layered atop biology; it is a practical generative language that encodes constraints, promotes reuse across scales, and yields tangible candidates for de novo proteins.

Viewed through the lens of deep time, Deep Aria is not simply a one-to-one translation but a transformation that reveals residual memory across scales. The Goldberg Aria’s counterpoint provides a scaffold whose encoded relationships, once mapped and folded, yield sequences that are musically cognate yet materially unprecedented, realizing an audible residue of structural memory that evolution itself never sampled. In this sense, folding introduces hidden information: the sequence-to-structure mapping adds new constraints and symmetries that the ear can apprehend as thematic recall, while the lab readout confirms a novel, realizable form.  

\subsection{Spider Web Sonification}

Spider webs provide a paradigmatic example of how structural hierarchies can be translated into the sonic domain. At the nanoscale, silk proteins assemble into fibrils and threads of remarkable toughness; at the macroscale, these threads form intricate, three-dimensional networks whose geometry encodes both mechanical resilience and ecological function. By digitizing full 3D web architectures through high-resolution imaging and reconstruction, we were able to analyze their topology as scale-free networks and, crucially, to sonify them \cite{Su2018JRSi}. Each fiber was mapped to a vibrating string, enabling the web to be experienced as a musical instrument whose timbre and pitch reflect underlying material properties \cite{Su2021CMJ}.  

Through collaborations with Tomás Saraceno and others~\cite{Saraceno2021}, these sonifications have been presented as interactive performances and installations, where audiences can literally listen to the architecture of a web and explore its structural memory through sound~\cite{Su2022JMUI}. The performances reveal how subtle differences in web construction yield distinct sonic identities, effectively allowing each spider to “compose” a unique musical piece. This mapping is not only aesthetic but also analytical: stretching or breaking threads alters pitch, making mechanical processes audible and enabling a novel form of scientific instrumentation.  
In this sense, the spider becomes a co-composer. Its behavior defines the score’s topology, while human performers translate that structure into sound. The collaboration blurs authorship: the spider writes through silk and vibration, the human interprets through resonance and gesture. Each inhabits a complementary part of a shared generative process, a sort of ``embodied duet'' across species.

In this way, spider web sonification illustrates the central premise of materiomusic: that vibrational information embedded in matter can be transposed into a musical domain where it becomes perceptible, interpretable, and even performable. It highlights the dual function of materiomusic as a scientific tool for probing structure and as a cultural medium for experiencing the hidden architectures of the natural world.

\subsection{Cross-Material Mappings}

Cross-material mappings extend the materiomusic program beyond biomolecules and webs to media such as water and fire, using vibration as the common currency, diving a bit deeper into the ideas introduced in Fig.~\ref{fig:vibration_generative}. In water-driven cymatics, molecular spectra are ``liquified'': protein normal modes are translated into acoustic drives that excite thin water films, producing macroscopic surface-wave patterns with rich spatial symmetries \cite{Buehler2020NanoFutures}. Remarkably, these wavefields preserve information about sequence and fold class at human-observable scales; trained convolutional networks can classify proteins from water-surface images alone and even reveal folding state or binding events. The same pipeline can be inverted to generatively synthesize new images from learned features, demonstrating that a physical wave medium, coupled to learning, can serve as a tangible analogue computer for molecular structure. Here, water is not merely a display but rather it becomes a generative substrate that enforces the constraints of the wave equation, biasing representations toward physically admissible structure.

Fire offers a complementary testbed where flow, heat release, and acoustics entwine. The spatiotemporal flicker of flames carries a latent harmonic organization that can be sonified, turning “silent” flames into audible patterns and enabling cross-modal translation between sound and flame imagery \cite{Milazzo2021iScience}. By learning a bidirectional map between acoustic inputs, spectrotemporal representations, and camera frames of a candle flame, one can both listen to flame dynamics and compose soundscapes that sculpt target flame morphologies. Embedding this within a generative model (e.g., VAE-based latent spaces) yields a design loop: capture dynamics, learn a compact representation, traverse latent directions to specify new patterns, and materialize them, all the way up to and including 3D-printed artifacts whose geometry is informed by the learned ``grammar'' of fire. In this setting, fire acts as an embodied generator whose physics constrains and regularizes the space of possibilities.

\begin{figure}[h!]
\centering
\includegraphics[width=1\textwidth]{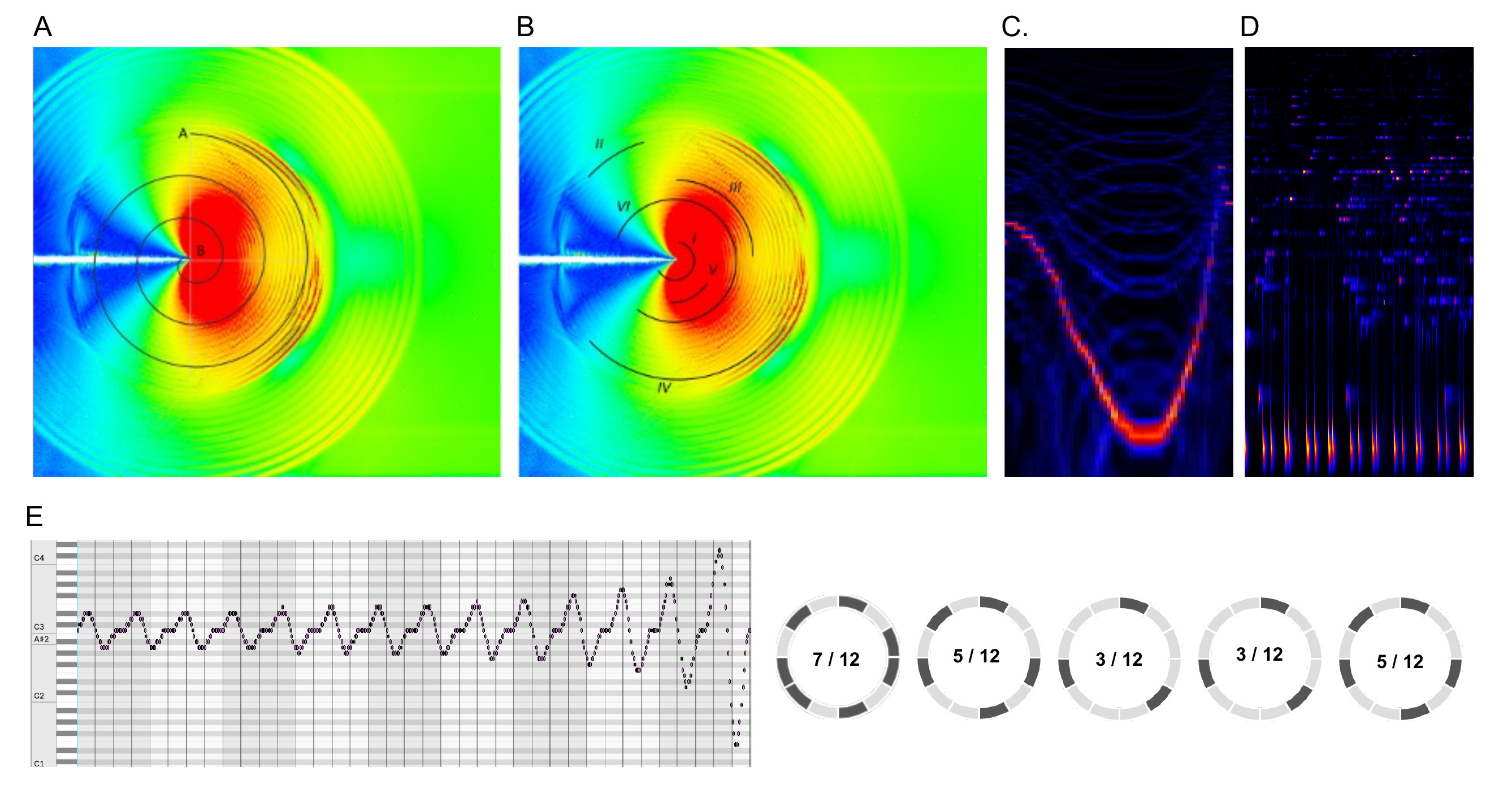}
\caption{Stress-field distribution near a crack tip translated into musical structures. (A) Near-tip principal stress field, highlighting tensile (red) and compressive (blue) lobes, mapped to continuous pitch modulation along a string. (B) Angular decomposition of the stress field into six sectors (I–VI), each encoding distinct directional stress variations that function as generative melodic fragments. These serve as templates for trajectories where tensile regions correspond to ascending motion, compressive regions to descending motion, and asymmetries to rhythmic subdivision. (C) Spectrogram of continuous pitch modulation derived from the field in (A). (D) Spectrogram of a melodic sequence generated from sectoral decomposition as in (B). (E) Example of a score derived from fracture-field mappings, notated on a pitch grid (left), and enumeration of scale structures (right), shown as circular diagrams indicating note occupancy (e.g., 7/12, 5/12), which link fracture-derived motifs to harmonic content. Together, the panels demonstrate how fracture mechanics fields provide both physical evidence of material failure and a latent musical grammar for composition and counterpoint.}
\label{fig:fracture_field_score}
\end{figure}

It is often said that music lives at the edge of chaos. This work asks a different question: what happens when music emerges from collapse itself? To probe this we developed a method for fracture sonification, where stress-tensor dynamics are rendered as sound at the moment of material singularity - an audible trace of deep time written in an instant of failure, linking atomistic vibration to macroscopic rupture. 
In the piece titled ``Counterpoint Through Fracture: Spiral Roosters'' (part of a series of compositions derived from the mathematics of rupture) we build on the universal geometry of a crack, it investigates whether the physical laws that govern material failure can also generate structure, coherence, and expression. Using the stress fields that form near the tip of a fracture. This is a region where tension approaches singularity at the scale of atoms. The piece utilizes these extreme conditions to construct a sonic architecture in which counterpoint arises not from convention, but from shared physical logic. It asks whether the deep order hidden within collapse might have its own music; notably not imposed, but revealed (Fig.~\ref{fig:fracture_field_score}). In Fig.~\ref{fig:fracture_field_score}A, diffraction imaging reveals the characteristic dipolar distribution of tension and compression radiating from the crack tip, a canonical pattern predicted by fracture mechanics. In Fig.~\ref{fig:fracture_field_score}B, the field is divided into six angular regions (I–VI), each representing a distinct zone of stress variation. These regions were treated as latent voices. To compose, trajectories were drawn through each sector, spiraling toward and away from the singularity. Local gradients in stress intensity were mapped to melodic features: sharp gradients produced rapid registral leaps and denser rhythms, while smoother regions yielded sustained tones and elongated gestures. Tension was translated into upward motion, compression into downward. Temporal structure was organized through sparse, asymmetric Euclidean subdivisions (3/12, 5/12, 7/12), whose irregularity reflects the asymmetry of material order (Panel E). 

In one example piece generated through this process, ``Counterpoint Through Fracture'', we choose to snap all produced notes to the \emph{C harmonic minor} scale, obtained by raising the seventh degree of the natural minor to yield the sequence C--D--E$\flat$--F--G--A$\flat$--B--C. This construction ensures a stable tonal framework: the raised leading tone (B) introduces functional dominant harmony (V and V$^7$), enabling authentic cadences and strong resolution to the tonic (here, C). The coexistence of A$\flat$ and B expands available contrapuntal resources, supporting smooth voice-leading and chromatic inflection, while the characteristic augmented second (A$\flat$--B) creates heightened melodic tension that, when distributed across independent voices, reinforces cadential strength and contrapuntal coherence.
Within this framework, the composition uncovered melodic lines by traversing the field. Counterpoint emerged as an intrinsic property of fracture geometry when multiple generative agents jointly produce notes.
This approach produces a polyphonic texture that is dense yet coherent. Each line preserves its independence, but their relationships are bound by the silent common ground of the stress field. The result is a form of field-generated counterpoint, where musical structure is not superimposed on material behavior but is instead revealed through it. The piece thus demonstrates how systems at the edge of failure - governed by singular fields and local asymmetries - can generate resilient coherence, transforming disintegration into construction.

This compositional method also gestures toward broader generative paradigms. Akin to the approach by which intelligent systems recombine fragments of learned distributions to synthesize novelty, or proteins mutate to create new functions, the fracture field operates here as a universal prior. Melodies are drawn from its structure rather than imposed upon it, and the assembly of fragments into a whole reflects the recombinatory logic of physical and biological systems, forming a physics-aware compositional logic. 
The work aligns with a lineage of composers who have sought structure through systems, from Xenakis through statistical mechanics, Nancarrow through automata, Reich through phase patterns, Grisey through spectral analysis. Yet unlike many algorithmic strategies, the gestures exposed here are grounded directly in physical reality and the counterpoint is not simulated but sensed. String vibrations are modeled as resonant bodies under stress, modulated by field variation, with rhythm generated by partitioned angular trajectories. The architecture of the composition is therefore audibly tangible, shaped by the same universal principles that govern rupture.
This piece illustrates how fracture, far from being a mere endpoint of material integrity, can act as a generative principle. By rendering the stress tensor audible, it transforms collapse into a source of coherence. The result is not music about fracture, but music composed from fracture and within that context, demonstrating how unstable systems can give rise to resilient forms, and how the mathematics of rupture can serve as an instrument of composition at the edge of material stability. 

These studies point to a broader principle. Wave media (strings, membranes, fluids, flames) naturally enforce conservation laws and boundary conditions that are difficult to impose in purely symbolic generative models. When we map structure into such media and back, we inherit their inductive biases: continuity, symmetry, resonance, and dispersion. The result is a form of physically grounded creativity where the medium co-authors the outcome. In this sense, wave media are not passive canvases but genuine co-authors. Strings vibrate only within the limits of their tension, water surfaces organize into standing waves constrained by the wave equation, and flames flicker in patterns dictated by thermofluid dynamics. By embedding these conservation laws directly into the generative process, the medium itself participates in invention, steering outcomes toward realizable, stable forms that could not emerge from symbolic computation alone.
Cross-material mappings thus serve two roles at once: they expose hidden invariants (what persists when translating between matter and sound), and they provide practical machinery for discovery (how to steer generative processes toward realizable forms). In water and fire, we see that generative algorithms do not live solely in code; they can be instantiated in matter itself, with waves as the scaffolding that carries ideas across domains \cite{Buehler2020NanoFutures,Milazzo2021iScience,Lu2023PNAS}.

Fire situates deep time at the scale of human evolution: a primordial medium through which nutrition, sociality, and threat were encoded into our collective memory. Sonifying flames by mapping characteristic flicker into harmonic spectra and back makes those latent patterns audible and composable, while the learned ``circle of fire'' reveals bidirectional control between acoustic harmonics and flame morphology. In doing so, the work surfaces evolutionary memory as a creative constraint: the medium’s physics regularizes what can be generated and steers invention toward forms that remain legible to human perception across vast temporal depths.   

\begin{figure}[ht]
\centering
\includegraphics[width=1.\textwidth]{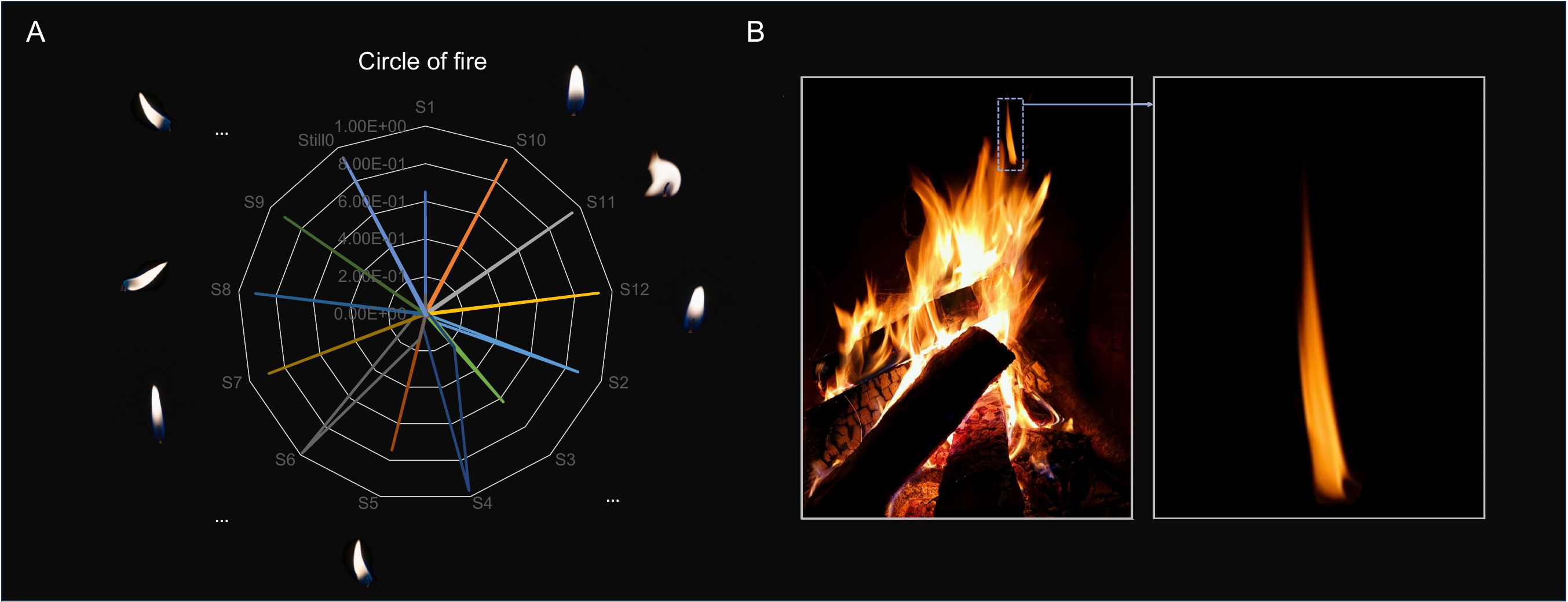}
\caption{
Circle of Fire as introduced in~\cite{Milazzo2021iScience}. Panel A: Radar-plot representation of flame morphologies mapped into a latent harmonic space. 
Each spoke corresponds to a learned spectral–temporal mode of flame flicker, while the surrounding 
images illustrate representative states of the flame. The diagram highlights the bidirectional mapping 
between sound and fire: harmonic inputs sculpt target flame shapes, and flame dynamics can be read 
as musical spectra. This exemplifies how wave media act as co-authors in materiomusic, embedding 
physical laws of fluid and thermal dynamics directly into generative processes. Panel B shows the compositional breakdown of a larger fire into flame components.
}
\label{fig:circle_of_fire}
\end{figure}

As shown in Fig.~\ref{fig:circle_of_fire}, these mappings can be organized into a ``circle of fire,'' 
where distinct morphologies of a flame correspond to latent harmonic modes. 
This representation highlights the bidirectional nature of the approach: 
acoustic inputs can sculpt target flame shapes, while the spatiotemporal 
flicker of the flame can be translated into harmonic spectra. In this way, 
fire becomes both a medium of sonification and a generator of new 
structural patterns, exemplifying the broader principle of wave media as 
co-authors in materiomusic.

\section{Discussion and Outlook}

Across the examples discussed here, a common mechanism emerges that reframes novelty in physical, biological, and cultural systems. Rather than arising from randomness, recombination, or the introduction of new laws, novelty appears when a system is forced to reorganize its own space of viable configurations because existing degrees of freedom are insufficient to satisfy imposed constraints. This view builds on early insights that form is shaped by physical law rather than blueprint \cite{thompson1917growth}, and that new macroscopic patterns can emerge when a previously stable regime becomes unstable under unchanged governing dynamics \cite{turing1952morphogenesis}. Here, however, the emphasis is on constraint incompatibility as the generative trigger: when mechanical, energetic, chemical, or informational constraints cannot be jointly satisfied within a given representational space, persistence requires an effective enlargement of that space through the appearance of new modes, new variables, or new structural operators. In this sense, selective imperfection is not an aesthetic preference but the stationary regime of systems that inevitably generate variation while filtering it through constraints that preserve coherence and adaptability. This interpretation resonates with Prigogine’s account of dissipative structures stabilized far from equilibrium by the interplay of fluctuation and constraint \cite{prigogine1980being}, and with Bergson’s insistence that time entails genuine emergence rather than mere rearrangement \cite{bergson1911creative}, while remaining mechanistic and testable. It also recasts generative syntax in physical terms: whereas Chomsky formalized syntax as a rule system enabling unbounded structural generation \cite{chomsky1957syntactic}, the present framework treats syntax as the set of constraint-respecting transformations that keep a system realizable under dynamical law, with novelty corresponding to the emergence of new permissible transformations when old ones fail. In effect, we propose a thermodynamic definition of creativity. 
Concrete instances of this mechanism include fracture, where a single smooth crack can no longer satisfy energy release and stability constraints under increasing load and the system reorganizes by introducing branching and roughness as new effective degrees of freedom; fluid flow, where laminar solutions cease to transport momentum efficiently and vortical or turbulent structures emerge; morphogenesis, where homogeneous states give way to patterned modes; and biological evolution, where variation and selection stabilize new motifs, domains, and regulatory couplings that expand the space of foldable and functional configurations \cite{kauffman1993origins,holland1975adaptation}. Living systems make this distinction explicit: life corresponds to sustained closure, in which constraint incompatibilities are repeatedly resolved through reorganization, whereas death marks an absorbing boundary where such reorganization can no longer maintain viability; yet even death can become a constructive operator when rendered selective and rule-governed, as in programmed cell death sculpting multicellular form \cite{kerr1972apoptosis}. Viewed this way, novelty is not the appearance of new forms per se, but the forced enlargement of the space of viable forms under constraint, with new forms arising as stable realizations within this expanded space. This is a process that operates across scales and domains, and that iterative generative procedures such as diffusion models and swarm dynamics approximate computationally by repeatedly introducing variation and restoring constraint satisfaction.

In this view, art and science are isomorphic at their core, not because they share methods or goals, but because they enact the same generative mechanism. Both arise from sustained engagement with constraint (physical, formal, cognitive, or cultural), and notably, both produce novelty by enlarging the space of viable forms when existing degrees of freedom can no longer satisfy those constraints. New artistic forms and new scientific constructs thus emerge not as arbitrary inventions, but as stable realizations within an expanded space of possibility forced into existence by constraint incompatibility. This shared structure provides a unifying mapping between creative practice and scientific discovery, suggesting that art and science are complementary modes of probing and extending the same underlying generative logic by which nature itself organizes, adapts, and invents. 
Examples of this mechanism span art and science alike, from the emergence of polyphony when monophonic musical structures can no longer sustain complexity, to the introduction of perspective when flat depiction fails to capture spatial coherence, to crack branching under load when a single fracture path cannot dissipate energy, and to the appearance of new macroscopic order parameters when existing physical variables cease to describe stable behavior.

From a practical matter, the work described here positions materiomusic as a natural language of life, one that unites molecules, sound, and creativity under a shared grammar of vibration, hierarchy, and transformation. 
A key insight is that selective imperfection acts as the conceptual bridge between matter and music. 
In materials, intermediate defect densities enhance toughness, stability or strength (Fig.~\ref{fig:defects}), while in music, moderate irregularity enables expressive structure Fig.~\ref{fig:entropy_cultural_combo}. 
Although derived from distinct domains, both follow the same generative rule: richness emerges not from perfection or chaos, but from a balanced interplay of order and disorder (Fig.~\ref{fig:entropy_evenness_bridge}). 
This shared grammar underscores that materiomusic serves as a unifying design principle that links the physics of materials with the cultural evolution of scales and sound. 

Listening is hence not passive reception but participation. When we hear the vibrations of a spider web or a protein fold, we engage our own vibrational architecture (bones, membranes, and neural oscillations)as resonant media. The act of listening becomes a form of modeling: we reconstruct external structures within ourselves. Materiomusic thus operates not only as an external mapping between domains but as an embodied correspondence, where perceiver and phenomenon co-vibrate within a shared physical continuum.
Across proteins, spider webs, fire, and water (and likely many more systems), we find that the principles governing structural assembly are echoed in the organization of music. 
Both domains thrive on the interplay between universality and diversity, symmetry and imperfection, and local motifs and global form. 
This recognition elevates materiomusic from analogy to a functional design language, capable of producing new insight, guiding experimentation, and inspiring artistic expression \cite{Cranford2010NSA}.

\begin{figure}[ht]
\centering
\includegraphics[width=.6\textwidth]{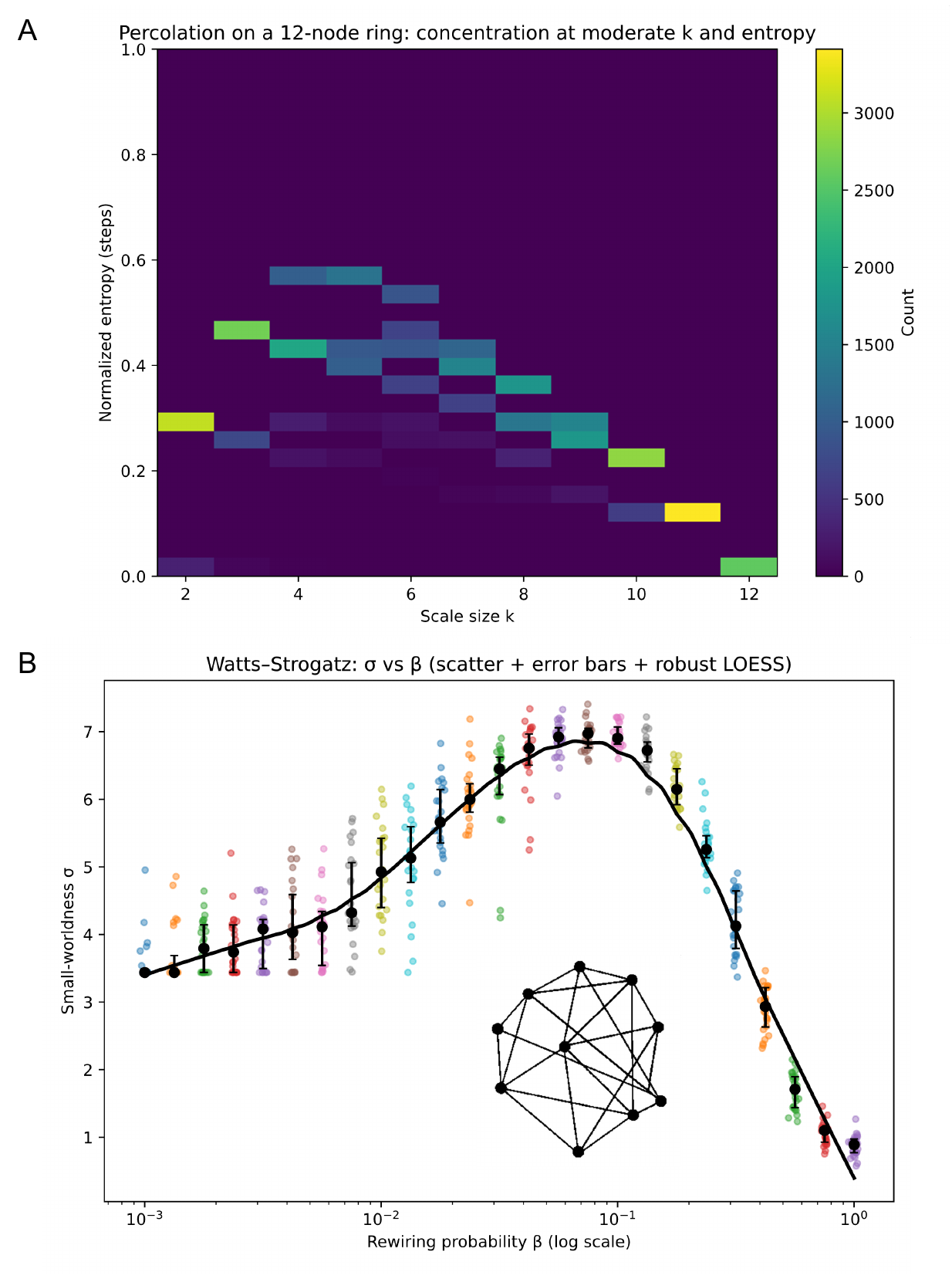}
\caption{Emergent structure at intermediate randomness across domains. 
(A) Random percolation on a 12-node ring (twelve pitch classes) generates musical scales whose distribution concentrates at moderate size ($k \approx 5$--8) and intermediate entropy ($H \approx 0.3$--0.5), matching the regime occupied by many cultural scales such as major, minor, and pentatonic. 
(B) In the Watts–Strogatz model ($n=120$, $k=6$), small-worldness $\sigma$ peaks at intermediate rewiring probability $\beta$, where networks preserve local clustering while gaining efficient long-range shortcuts. 
Together, these analyses illustrate a common structural principle: both musical scales and complex networks become most functional and expressive not in states of perfect order or complete randomness, but in the balanced regime between them. This underscores a variety of points made in this paper, including small-worldedness of swarm-based compositions (Fig.~\ref{fig:swarm_music}) and the properties of cultural musical scales (Fig.~\ref{fig:entropy_cultural_combo}).} 
\label{fig:emergence}
\end{figure}

 To clarify how balance between order and randomness shapes structure, we turned to two minimal computational models, shown in Fig.~\ref{fig:emergence} (details in Materials and Methods). The first model explores music: by randomly activating pitch classes on the twelve-tone circle, we simulate the spontaneous formation of scales. If such activations were uniformly random, we would expect scales to scatter across all sizes and interval distributions. Instead, as Fig.~\ref{fig:emergence}A shows, the outcomes concentrate at intermediate scale size ($k \approx 4$--7) and moderate entropy ($H \approx 0.3$--0.5), which is also where many cultural scales, including major, minor, and pentatonic, are found. 

The second model explores networks using the Watts-Strogatz framework. Starting from a ring of $n=120$ nodes, each connected to $k=6$ nearest neighbors, we introduce randomness by “rewiring” edges: with probability $\beta$, a local link is redirected to a randomly chosen distant node. This interpolates smoothly between a perfectly ordered lattice ($\beta=0$) and a completely random graph ($\beta=1$). As shown in Fig.~\ref{fig:emergence}B, the small-worldness $\sigma$ peaks at intermediate $\beta$, a regime that preserves local clustering while introducing efficient long-range shortcuts. 

Viewed together, the two experiments illustrate the same law from different domains. In the musical percolation model (panel A), scales concentrate in the middle ground between rigid order and chaotic irregularity. In the network model (panel B), small-worldness likewise emerges at an intermediate balance of order and randomness. In both cases the extremes are unproductiv. We see that pure order is rigid and limited, pure randomness incoherent, while the intermediate zone produces the richest and most functional structures. This parallel underscores that the cultural emergence of common scales reflects the same universal design principle that governs the architecture of complex networks.

Beyond these cultural instantiations, materiomusic is not simply an act of 
sonification but of functional modeling. The mappings between matter 
and music are designed to preserve categories, hierarchies, and relationships 
such that listening itself becomes a mode of classification and inference. 
Protein folds, web modules, or flame morphologies are not merely made audible, 
they are translated into perceptual grammars where differences in pitch, rhythm, 
or timbre correspond to differences in physical function or structure. In this 
sense, materiomusic is not an after-the-fact soundtrack but a modeling tool: it 
projects functional categories into the auditory domain, allowing researchers 
and audiences alike to hear distinctions that matter scientifically. The aim is 
to construct reversible, physics-grounded correspondences where sound does not 
just represent, but actively models the functional logic of the material 
world, providing the dynamics of matter as co-author in the creative process.

The examples and arguments presented here suggest that materiomusic is more than an analogy: it is a generative language that unites molecules, sound, and creativity through shared principles of vibration, hierarchy, and imperfection. By rendering hidden architectures audible and by embedding musical structures into material design, this framework enables new ways of probing scientific data, inventing resilient materials, and creating cultural artifacts. The integration of swarm-based, agentic AI further demonstrates that collective dynamics can generate outputs with structural coherence and novelty approaching human-like structural signatures of creativity. This may open a future in which discovery is conducted not only by analysis but by composition. In this sense, materiomusic offers both a microscope and a score: a way of perceiving what is otherwise invisible, and a method for constructing what does not yet exist. Embracing this dual role opens a path to a science that is inherently creative, weaving together matter, life, and art in a shared grammar of invention.

Critically, we posit that selective imperfection thus emerges not as an anomaly but as a stable regime of systems that inevitably generate variation while filtering it through constraints that preserve coherence and adaptability. In this sense, novelty is neither imposed nor optimized, but arises naturally as the steady-state outcome of constrained generative dynamics.

We summarize key concepts: 

\begin{tcolorbox}[
  colback=gray!5!white,
  colframe=white!60!black,
  title={\textsf{Generative Principles of Materiomusic}},
  fonttitle=\bfseries\sffamily,
  coltitle=black,
  left=2mm, right=2mm, top=1mm, bottom=1mm,
  boxsep=2pt
]
\small\sffamily

\textbf{1. Vibration as a Universal Generative Medium.} 
Vibration is treated as a physically grounded substrate that unites matter, sound, and cognition. 
Across molecules, webs, and musical systems, oscillatory dynamics encode structure and function through shared constraints of resonance, tension, and boundary conditions. 
This establishes a common computational grammar of nature in which both material and cognitive organization emerge from wave-based interactions.

\medskip
\textbf{2. Selective Imperfection as a Generative Algorithm.} 
Deviations from perfect symmetry or periodicity act not as defects but as mechanisms for exploration and adaptation. 
In materials, moderate defect densities enhance resilience; in music, asymmetry enables tension and narrative flow. 
Selective imperfection thus defines a universal algorithm for creativity, thereby perturbing equilibrium to open new regions of the design space.

\medskip
\textbf{3. Cross-Species and Cross-Scale Cognition through Vibration.} 
By mapping the vibrational universe of non-human systems (e.g., spiders) into the human auditory domain, 
materiomusic provides an interpretable interface between distinct perceptual worlds. 
This approach enables a form of vibrational epistemology, where perception is modeled as resonance rather than symbolic representation.

\medskip
\textbf{4. Generative Science as Composition.} 
Scientific modeling and musical composition are treated as equivalent generative operations. 
Bidirectional mappings between matter and sound convert discovery into a performative, iterative process in which physical law and human intention co-produce structure. 
Each translation from material dynamics to music and back generates new, testable information.

\medskip
\textbf{5. Incompleteness and Creativity.} 
Extending Gödel’s notion of incompleteness to physical and cognitive systems, 
materiomusic frames creativity as the natural consequence of operating at the boundary of formal closure. 
By embracing asymmetry and open-endedness, both material systems and intelligent agents transcend the limits of fixed rule sets, 
transforming imperfection into a driver of emergence and innovation.

\medskip
\textbf{Summary:} 
Together these principles outline a \emph{physics of generativity}: 
vibration provides the medium, selective imperfection supplies the algorithm, and cognition (human or artificial) emerges as resonance within this continuum. 
Materiomusic therefore functions not as a metaphor but as a scientific framework for understanding how matter, mind, and music co-generate form.

\end{tcolorbox}

The implications for science could be profound. Sonification and cross-modal mappings provide a new way to probe data, particularly in systems where visual inspection obscures complexity. Listening to proteins or webs reveals patterns of hierarchy, closure, and variation that conventional analytics may miss. At the same time, embedding these mappings into generative workflows has yielded de novo proteins with verifiable structural stability \cite{Yu2019ACSNano,Qin2019EML}. Similarly, cross-material explorations in water and fire demonstrate that wave-based media can function as generative substrates, embodying physical constraints that help regularize deep learning models and bias them toward realizable outcomes \cite{Buehler2020NanoFutures,Milazzo2021iScience}. In this way, materiomusic becomes a tool for discovery, not only communicating known truths but also shaping the space of the possible.

For engineering, materiomusic suggests a paradigm where design is guided by the same principles that make music compelling. Protein folds and spider silk architectures can be understood as compositions, balancing repetition and surprise, modularity and integration. This provides a powerful heuristic for designing new materials, particularly in the context of AI-driven workflows that aim to explore vast design spaces. Music-inspired design rules, such as hierarchical reuse, long-range memory, and the embrace of imperfection, can accelerate the creation of resilient, multifunctional, and adaptive materials \cite{Lu2023PNAS}. By grounding engineering practice in a generative, compositional framework, materiomusic expands the design space beyond conventional optimization into one infused with creativity.

Culturally, materiomusic bridges art and science in ways that foster new dialogues. The performances and installations of spider web sonifications, protein compositions, and antibody music exemplify how scientific structures can be experienced as art, and how artistic media can feed back into scientific exploration \cite{Su2021CMJ,Su2022JMUI}. Such work makes hidden architectures perceptible, rendering complex data emotionally legible. In doing so, it creates shared cultural artifacts that engage broad audiences in scientific ideas, blurring boundaries between laboratory, concert hall, and museum. This speaks to a deeper truth: science and art are not parallel domains but co-creative enterprises, each extending the reach of human imagination.
We note that where Kepler sought harmony in planetary motion, materiomusic finds it in the recursive interplay of matter, sound, and intelligence. 

Materiomusic also functions as a cultural instrument. By transforming proteins, webs, or flames into shared artifacts of sound, it dissolves the boundary between laboratory and stage. Gallery-scale web instruments, protein scores, and flame choirs render complex structure audible, making data emotionally legible and inviting public participation in discovery. These installations create a dual value: they serve as instruments for scientific inquiry while simultaneously acting as cultural objects that engage broad audiences in the hidden architectures of matter.

Looking forward, the framework of materiomusic may provide a sort of roadmap for a generative science of creativity itself. The swarm-based AI frameworks we have begun to develop show that collective dynamics can produce music with long-range coherence and structural memory that rivals human composition \cite{Buehler2025MusicSwarm}. These models suggest that novelty is not achieved by scaling monolithic learners but by embracing distributed agency, feedback, and emergent specialization. These are principles long familiar in both biology and music. More broadly, the fusion of sonification, generative AI, and experimental realization (from proteins expressed in the lab to additively manufactured webs) points to an integrated cycle of design where matter, sound, and intelligence coevolve.
Seen through this lens, composition becomes a method of remembering: an unconscious act that makes the past conscious by surfacing structural residues that persist across molecular, biological, and cultural epochs. Materiomusic thus operates as a deep-time instrument that allows us listening to matter to reveal ancient memory, and composing with that memory to construct new forms. 

Although this article has emphasized proteins, spider webs, water, and fire, the scope of materiomusic is broader. Any system in which vibration encodes structure is a candidate: membranes, lattice frameworks, even urban traffic flows understood as oscillatory fields. The unifying requirement is that mappings preserve relational information and respect the physics of the medium. This broader horizon suggests that materiomusic can serve as a general grammar for interpreting and designing complex systems across domains.

The key terms summarized in Table~\ref{tab:glossary} highlight the shared 
grammar that underpins materiomusic. Hierarchy describes a nested organization 
across scales in both matter and music. Sonification denotes the translation of 
data into sound, while isomorphism ensures that mappings between domains preserve 
structure and enable reversibility. Transpositional equivalence allows molecular 
vibrations to be shifted into the auditory range without losing their internal 
logic. The Hall--Petch effect illustrates how selective imperfection can enhance 
material strength, just as uneven scales create musical expressivity. Network 
concepts such as small-worldness, modularity, and participation coefficients 
capture how motifs cluster and recur across sections. Stigmergy explains the 
indirect coordination that drives swarm-based composition. De novo design refers 
to the creation of entirely new sequences, and cymatics shows how vibration 
generates visible patterns in physical media. These concepts articulate 
a unified vocabulary that binds matter, music, and intelligence.
The interface of machine and discovery, and scientific progress, is particularly compelling. Future discovery increasingly emerges from human–machine symbiosis, where machines are not merely instruments but co-creators of insight. This framing resonates with our approach to vibration as both structure and computation in matter and music~\cite{Wang2025Prizes}.

The closing vision is that novelty arises not from invention \textit{ex nihilo}, but from the recursive interplay of structure, imperfection, and collective dynamics. In this view, materiomusic is not a curiosity at the margins of art and science but uncovers a prototype of how discovery itself may evolve in the era of AI. By listening to matter and composing with its forms, we begin to shape a new generative language for creativity that transcends disciplinary boundaries and reflects the deep unity of life, sound, and structure. Novelty arises not from data alone but from the music of matter itself - it serves as as a universal language waiting to be composed.  
Ultimately, materiomusic asks a generative question rather than a descriptive one: not what exists, but why and how structures come into being. It seeks the meta-algorithms that generate form across domains, from the spider weaving its web to the human composing and performing a song. As we trace these processes through vibration, we approach a science of creation at its core, and is hence best understood as an inquiry into the mechanisms by which matter becomes music and the world composes itself.

\begin{table}[ht]
\centering
\renewcommand{\arraystretch}{1.2}
\small 
\begin{tabular}{p{0.22\textwidth} p{0.70\textwidth}}
\hline
\textbf{Term} & \textbf{Definition} \\
\hline
Hierarchy & Nested organization across scales. In matter: atoms $\to$ molecules $\to$ fibers $\to$ tissues; in music: notes $\to$ motifs $\to$ phrases $\to$ full forms. \\

Sonification & Translation of data into sound to reveal patterns. In materiomusic, mappings aim to preserve structural relationships rather than act as superficial overlays. \\

Isomorphism & A structure-preserving map between domains. Ensures that relationships in matter correspond to relationships in music. \\

Transpositional Equivalence & Shifting all pitches by the same amount preserves relative structure, allowing molecular vibrations to be shifted into the human auditory range. \\

12-TET & Twelve-tone equal temperament, the dominant tuning system in Western music dividing the octave into 12 equal steps. Different analyses in this paper use distinct constraints on what denotes a valid scale.\\

Hall-Petch Effect & In crystalline materials, reducing grain size increases strength up to a point, after which the inverse Hall--Petch regime causes softening. \\

Small-Worldness & A network property combining local clustering with efficient long-range connections. In music, corresponds to motifs that recur across time to create global coherence. \\

Modularity & The degree to which a network divides into distinct communities. In music, analogous to sections such as verse, chorus, or bridge. \\

Participation Coefficient & A measure of how much a node (e.g., a motif) connects across communities versus remaining isolated. Indicates thematic integration. \\

Stigmergy & Indirect coordination via shared traces (e.g., pheromone trails or musical motifs), enabling distributed systems to self-organize without central control. \\

\textit{De Novo} Design & Creation of something entirely new from first principles, e.g., composing novel protein sequences with stable folds rather than modifying existing ones. \\

Cymatics & Visualization of vibration through standing wave patterns in a medium such as water or sand. Connects vibrational signatures to visible geometry. \\

Co-Authoring Media & Wave media (strings, water, flames) that enforce conservation laws and boundary conditions, acting as active participants in generative processes rather than passive canvases. \\

Epistemic Inversion & The use of listening as a scientific instrument, where auditory perception becomes a means of probing categories and hierarchies usually studied visually. \\

Cultural Instrumentality & The role of materiomusic in dissolving boundaries between lab and stage, producing shared artifacts that make scientific data emotionally legible and invite public participation. \\
\hline
 \\
\end{tabular}
\caption{Glossary of key terms in materiomusic, extended with concepts from cultural practice and physics-grounded co-authorship, linking materials science, music theory, and network analysis.}
\label{tab:glossary}
\end{table}

\section{Materials and Methods}

\subsection{Stress-field distribution near a crack tip as note generator}

A Max for Live device is used to generate MIDI events by mapping fracture mechanics fields into musical pitch space. The input parameters are the polar angle $\theta$, radius $r$, and the stress intensity factor (SIF). In the selected mode (stress measure = 2), the governing relation implemented in the patch is
\[
\sigma^{\text{patch}}_1(\theta,r) \;=\; \frac{\mathrm{SIF}}{\sqrt{r}} \,\big(1 - \sin\theta \big)\cos\theta ,
\]
which serves as a Mode~I–like approximation of the near-tip principal stress field. This form retains the characteristic $1/\sqrt{r}$ singularity and alternation between tensile ($\sigma_1>0$) and compressive ($\sigma_1<0$) lobes as $\theta$ varies, but simplifies the canonical half-angle dependence into a two-lobe envelope that is particularly well suited for sonification. The normalization factor is relative and mapped to MIDI ranges, so absolute scaling (e.g.\ $1/\sqrt{2\pi}$) is absorbed into the musical mapping. 

The device increments $\theta$ and $r$ in user-defined steps ($\Delta\theta$, $\Delta r$), scales the output by the SIF, and maps the resulting values onto MIDI note numbers with optional transposition. Musical scaling can be incorporated by snapping all produced notes to a scale, e.g., C harmonic minor. A playback toggle selects between monophonic (only the new note played) or polyphonic (new and old notes layered) behavior. The evolving sequence is displayed in a rolling table and sent to Ableton Live instruments, enabling direct sonification of crack-tip stress singularities as structured pitch trajectories.

\subsection{Defect and structure analysis in musical scales}

The analysis is inspired by~\cite{AllTheScales,ZeitlerAllScales}.  
All $2^{12}$ possible subsets of the twelve pitch classes in equal temperament
were exhaustively enumerated. A subset was retained as a valid scale if it
contained the root (pitch class 0). For panels where a maximum circular step
constraint was applied (Fig.~\ref{fig:defects}B), only scales in which no interval
between adjacent notes on the pitch-class circle exceeded the chosen threshold
(3, 4, or 5 semitones) were kept. Enumeration was implemented in Python using
bitmask representations of subsets.
 
For each valid scale with $k$ notes, we derived the step vector
$\mathbf{g}=(g_1,\dots,g_k)$ by computing semitone distances between consecutive
notes around the circle (including wrap-around). The evenness defect was defined
as the standard deviation of $\mathbf{g}$ relative to the uniform step length
$12/k$, normalized to the range [0,1] by dividing by the maximum standard
deviation observed at that $k$. This yields 0 for perfectly even scales and 1
for the most uneven scales at that size. Fig.~\ref{fig:defects}C shows the
distribution of evenness defect values binned into 20 equal-width bins between
0 and 1 for each $k$.

  We quantify unevenness by the minimal
  number of semitone ``unit moves'' required to transform a scale’s
  interval vector into the nearest perfectly even partition of the
  octave. For a scale with $k$ notes and step vector $g = (g_1,
  \dots,g_k)$ satisfying $\sum_i g_i = 12$, we construct the ideal
  integer pattern with $\lfloor 12/k \rfloor$ and $\lceil 12/k
  \rceil$ steps so that it also sums to 12. We then sort both $g$ and
  the ideal pattern, compute their $\ell_1$ distance, and divide by
  two; each $+1/-1$ adjustment contributes $2$ to the $\ell_1$ gap,
  so $\tfrac{1}{2}\|g_{\text{sorted}} - g_{\text{ideal}}\|_1$ counts
  the fewest semitone shifts needed to reach even spacing. This
  integer ``unevenness defect count'' is zero exactly when the scale is
  already evenly spaced.

Counts of valid scales (Fig.~\ref{fig:defects}B) and distributions of evenness
defect (Fig.~\ref{fig:defects}C) were aggregated by number of notes $k$. All data shown were generated with Python scripts, which enumerates valid scales, applies
the max-gap constraint when specified, computes step vectors, and exports
metrics in tabular form.

\label{defect_scales}

As an additional analysis, we follow the approach suggested in \cite{ZeitlerAllScales,RingScales} to measure the concept of defects in a scale. Here we modeled the twelve-tone equal-tempered chromatic set as the pitch-class universe
\[
\{0,1,\dots,11\},
\]
where integers are understood modulo $12$. A candidate scale $S$ is represented as a bit mask over these 12 pitch classes; iterating over all integers $m$ from $1$ to $2^{12}-1$ enumerates all possible subsets.

We imposed three constraints to select musically plausible scales. First, pitch class $0$ (taken as the tonic) must be present in the scale (root constraint). Second, scales must contain at least three distinct pitch classes ($|S|\ge 3$; cardinality constraint). Third, writing the elements of $S$ in ascending order as $p_1 < p_2 < \dots < p_n$, we form the cyclic interval sequence
\[
d_i =
\begin{cases}
p_{i+1} - p_i, & i = 1,\dots,n-1,\\[2pt]
(p_1 + 12) - p_n, & i = n,
\end{cases}
\]
measured in semitones. A subset $S$ is accepted as a valid scale if all steps satisfy $d_i \leq 4$ for all $i$, i.e.\ no step is larger than a major third (maximum-step constraint).

For each valid scale $S$ we computed: (i) the scale size $n = |S|$, i.e.\ the number of notes in the scale; (ii) the number of \emph{imperfections} (defects); and (iii) the Shannon entropy of its interval pattern. For the defect count, we followed Zeitler’s definition: for each pitch class $p \in S$, we examined the perfect fifth above it, $(p+7) \bmod 12$. If this fifth was not contained in $S$, i.e.\ $(p+7) \bmod 12 \notin S$, then $p$ was counted as an imperfect degree. The defect number of $S$ is the total count of such imperfect degrees. For the entropy, we considered the multiset of interval sizes $\{d_i\}$, counted the frequency of each distinct step size, normalized to obtain probabilities $p_j$, and evaluated the Shannon entropy in bits as
\[
H(S) = -\sum_j p_j \log_2 p_j.
\]

We aggregated these quantities into a two-dimensional frequency table
\[
C(n,k) = \text{number of valid scales with } n \text{ notes and } k \text{ imperfections},
\]
for $n = 3,\dots,12$ and all defect values $k$ observed in the enumeration. Fig.~\ref{fig:scales-defects-entropy}A was generated by viewing this table as a function of the defect number: for each defect value $k$, we computed $C(n,k)$ for all $n$ and plotted a stacked bar at horizontal position $k$, whose segments represent the contributions from each scale size $n$ (3–12 notes). The height of the bar equals the total number of scales with $k$ imperfections, $\sum_{n} C(n,k)$, which is written as a numeric label above the bar. Fig.~\ref{fig:scales-defects-entropy}B was generated by transposing this view: for each scale size $n$ we computed $C(n,k)$ over all imperfection counts $k$ and plotted a stacked bar at horizontal position $n$. The height of each bar equals the total number of valid scales with $n$ notes, $\sum_{k} C(n,k)$, again annotated numerically above the bar. In both panels, different values of $n$ or $k$ are represented by distinct shades within a monochromatic blue palette, and the $y$-axis reports the total number of distinct scales satisfying the constraints above.

Fig.~\ref{fig:scales-defects-entropy}C summarizes the relationship between defects and interval-structure complexity. For each defect count $k$, we collected the entropies $H(S)$ over all scales with that number of imperfections, and computed the mean and population standard deviation,
\[
\bar{H}(k) = \frac{1}{N_k} \sum_{S:\,\text{defects}(S)=k} H(S), 
\qquad
\sigma_H(k) = \sqrt{\frac{1}{N_k}\sum_{S:\,\text{defects}(S)=k} \bigl(H(S)-\bar{H}(k)\bigr)^2},
\]
where $N_k$ is the number of such scales. We then plotted $\bar{H}(k)$ as a function of $k$, with vertical error bars indicating $\sigma_H(k)$.

\subsection{Entropy–Arrangement Scatter with Cultural Overlays}

We enumerated pitch-class sets (“scales”) on an $n$-tone equal-tempered circle ($n=12$ unless noted) subject to:
  (i) inclusion of the root class ($0$), (ii) optional maximum circular step constraint $g_{\max}\le G$ (here $G=4$), and
  (iii) size bounds $k\in[k_{\min},k_{\max}]$ (default $1\le k\le n$).
  For each scale (bitmask) we formed the sorted pitch classes $\{p_i\}$ and computed its step vector
  $g=(g_1,\dots,g_k)$ as circular differences with wrap-around, satisfying $\sum_{i=1}^k g_i = n$.

  Entropy was computed on the empirical distribution of step sizes in $g$~\cite{Shannon1948,Paninski2003,KasparSchuster1987,ZivLempel1978}.
  Let $p_s$ be the relative frequency of step size $s$ in $g$. The Shannon entropy (in bits) and its normalization are
  \[
  H=-\sum_{s} p_s \log_2 p_s,
  \qquad
  H_{\mathrm{norm}}=\frac{H}{\log_2(\min(k,n))}\in[0,1],
  \]
  which permits comparison across different $k$.

  Arrangement symmetry was quantified by an ``arrangement defect''
  \[
  A \;=\; 1 - \max_{\tau\in\{0,\dots,k-1\}}
  \frac{\langle g,\; \mathrm{rot}_\tau(\mathrm{rev}(g))\rangle}{\|g\|_2\,\|\mathrm{rev}(g)\|_2},
  \]
  equal to $0$ for step patterns that are palindromic under rotation, and increasing with asymmetry.

  The scatter plot places each enumerated scale at $(A,\,H_{\mathrm{norm}})$.
  Cultural exemplars (12-TET approximations) were overlaid using the same metrics:
  Western major/minor, major/minor pentatonic, diminished octatonic (whole–half and half–whole),
  harmonic minor (7), bebop dominant (8), bebop major (8), bebop harmonic minor (8),
  raga Bhairav, raga Kalyani (Lydian), maqam Bayati, and maqam Hijaz.
  To avoid overplotting, each exemplar uses a distinct color/marker and small, non-inferential jitter when coordinates coincide;
  a legend outside the axes identifies labels. 
  Scales that do not have a direct 12-TET analogue are effectively projected (quantized) onto the 12-TET grid (e.g., for overlays). The algorithm hence has no notion of ``true'' non-12-TET structure; it only sees their 12-TET approximations.

\subsection{Simple Computational Models of Emergent Structure}

We generated two simple complementary computational experiments to explore how balance between order and randomness leads to emergent structure in music-inspired systems and network theory. First, we implemented a site percolation model on a 12-node ring representing the twelve pitch classes of the chromatic octave. Each node was activated independently with probability $p$, sampled from $0.05 \leq p \leq 0.95$ in 19 steps, with $40{,}000$ realizations in total. If the tonic (pitch class 0) was not activated, it was forcibly included to ensure comparability. From each active set we derived a musical scale by sorting the active pitch classes and computing the corresponding circular step vector (interval sizes in semitones). We quantified the distribution of steps using a Shannon entropy measure, normalized to the theoretical maximum ($\log 11$), and recorded the scale size $k$, entropy $H$, and activation probability $p$. The resulting dataset was visualized as a two-dimensional histogram of scale size $k$ versus entropy $H$.

Second, we examined the small-world regime of networks using the Watts--Strogatz model~\cite{watts1998collective,PhysRevLett.87.198701}. 
We constructed graphs with $n=120$ nodes and mean degree $k=6$, and rewired edges with probability $\beta$ sampled logarithmically from $10^{-3}$ to $1$ (25 values). 
For each $\beta$, we generated 24 independent realizations. 
To establish a random-graph baseline, we computed the average clustering coefficient $C_{\mathrm{rand}}$ and characteristic path length $L_{\mathrm{rand}}$ for Erd\H{o}s--R\'enyi graphs with edge probability $p \approx k/(n-1)$, averaged over 8 replicates. 
For each Watts--Strogatz realization we measured the clustering coefficient $C$ and characteristic path length $L$, restricting to the largest connected component when necessary. 
Small-worldness was defined as
\[
\sigma \,=\, \frac{C/C_{\mathrm{rand}}}{L/L_{\mathrm{rand}}}\!,
\]
yielding a distribution of $\sigma$ values for each $\beta$. 
For visualization we plotted all trial results as scatter points (with small multiplicative jitter on the $\beta$ axis), overlaid per-$\beta$ medians with interquartile-range error bars, and fitted a smooth trend using robust LOESS in $\log_{10}\beta$ (tricube kernel with bisquare reweighting), evaluated across the full $\beta$ range. 
All simulations were implemented in Python using \texttt{NumPy}, \texttt{matplotlib}, and \texttt{NetworkX}, with fixed master seeds and independent trial seeds to ensure reproducibility.

\subsection{Size Effect Data Normalization}

For each dataset, values were normalized relative to the peak performance observed. The maximum performance value ($\mathrm{perf}^*$) and the corresponding length scale ($L^*$) were first identified. Performance values were then scaled as $y_{\mathrm{norm}} = y / \mathrm{perf}^*$, and the length axis was rescaled as $x_{\mathrm{norm}} = L^* / L$, inverse scaling. This procedure aligns all datasets at $x_{\mathrm{norm}}=1$, $y_{\mathrm{norm}}=1$, enabling comparison across systems with different absolute units and revealing universal features of size-effect trends.
We constructed median and envelope curves by binning normalized data on each side of the peak ($x = L^*/L$) into equal-weight logarithmic intervals, computing quantiles within each bin and interpolating them with a shape-preserving piecewise cubic Hermite interpolating polynomial (PCHIP). The shaded band corresponds to the inter-quantile range (e.g., 30--70\%), while the dashed curve shows the bin-wise median, both anchored at the normalized peak performance peak point $(1,1)$.

\subsection{Use of generative AI}
Generative AI was used in generating code and analysis. 

\section*{Supplementary Materials}
An audio summary of this paper is provided as Supplementary Material, generated using PDF2Audio~\url{https://huggingface.co/spaces/lamm-mit/PDF2Audio}.

\section*{Code and Data}
Codes and additional data are available at \url{https://github.com/lamm-mit/MusicAnalysis}. A dataset with scale defects and related properties is available at \url{https://huggingface.co/datasets/lamm-mit/scales-12tet-defects}.

\section*{Acknowledgments}
This work was supported in part by the MIT Generative AI Initiative.


\bibliographystyle{naturemag}
\bibliography{references
}

\end{document}